\documentclass[runningheads]{llncs}

\usepackage{eccv}

\usepackage{eccvabbrv}

\usepackage{graphicx}
\usepackage{booktabs}
\usepackage{multirow}
\usepackage{tikz-imagelabels}
\usepackage{graphicx}

\usepackage[accsupp]{axessibility}  

\usepackage{hyperref}

\usepackage{orcidlink}

\begin{document}

\title{Restoring Images in Adverse Weather Conditions via Histogram Transformer} 


\author{Shangquan Sun\inst{1,2}
\orcidlink{0000-0002-6292-2495} 
\and 
Wenqi Ren\inst{3,4}$^{\dag}$
\orcidlink{0000-0001-5481-653X} 
\and 
Xinwei Gao\inst{5} 
\and 
Rui Wang\inst{1,2}
\orcidlink{0000-0002-4792-1945} 
\and 
Xiaochun Cao\inst{3}
\orcidlink{0000-0001-7141-708X} 
}

\authorrunning{S.~Sun et al.}

\institute{Institute of Information Engineering, Chinese Academy of Sciences, Beijing 100085, China \and
School of Cyber Security, University of Chinese Academy of Sciences, Beijing 100049, China \and 
School of Cyber Science and Technology, Shenzhen Campus of Sun Yat-sen University, Shenzhen 518107, China \and
Guangdong Provincial Key Laboratory of Information Security Technology, Guangzhou 510006, China \and
Wechat Business Group, Tencent, Shenzhen, Guangdong, China, 518057 \\
\email{shangquansun@gmail.com},
\email{renwq3@mail.sysu.edu.cn},
}

\maketitle

\renewcommand{\thefootnote}{}
\footnotetext{\dag \ Corresponding author.}

\begin{abstract}
%
Transformer-based image restoration methods in adverse wea-ther have achieved significant progress.
Most of them use self-attention along the channel dimension or within spatially fixed-range blocks to reduce computational load. 
However, such a compromise results in limitations in capturing long-range spatial features.
Inspired by the observation that the weather-induced degradation factors mainly cause similar occlusion and brightness, in this work, we propose an efficient \textbf{Histo}gram Trans\textbf{former} (Histoformer) for restoring images affected by adverse weather. 
It is powered by a mechanism dubbed histogram self-attention, 
which sorts and segments spatial features into intensity-based bins.
Self-attention is then applied across bins or within each bin to selectively focus on spatial features of dynamic range and process similar degraded pixels of the long range together.
To boost histogram self-attention, we present a dynamic-range convolution enabling conventional convolution to conduct operation over similar pixels rather than neighbor pixels.
We also observe that the common pixel-wise losses neglect linear association and correlation between output and ground-truth. 
Thus, we propose to leverage the Pearson correlation coefficient as a loss function to enforce the recovered pixels following the identical order as ground-truth.
Extensive experiments demonstrate the efficacy and superiority of our proposed method.
We have released the codes in \href{https://github.com/sunshangquan/Histoformer}{Github}.
\end{abstract}
\abovedisplayshortskip=6pt
\belowdisplayshortskip=6pt
\abovedisplayskip=6pt
\belowdisplayskip=6pt

\begin{figure}[t]
\setlength{\abovecaptionskip}{5pt}
\setlength{\belowcaptionskip}{0pt}
  \centering
  \begin{minipage}{0.356\linewidth}
    \centering
    \begin{subfigure}{1\linewidth}
    \includegraphics[width=1\linewidth]{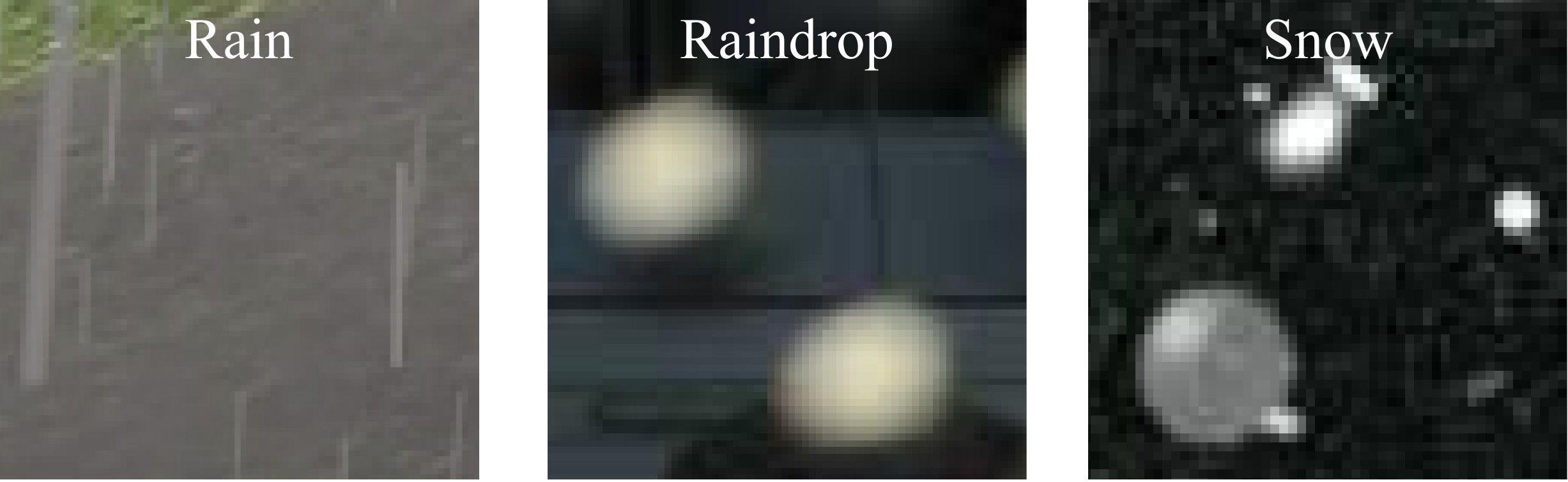}
    \subcaption[]{Input patches}
    \label{fig:intro-a}
    \end{subfigure}
    \hfill
     \begin{subfigure}{1\linewidth}
    \includegraphics[width=1\linewidth]{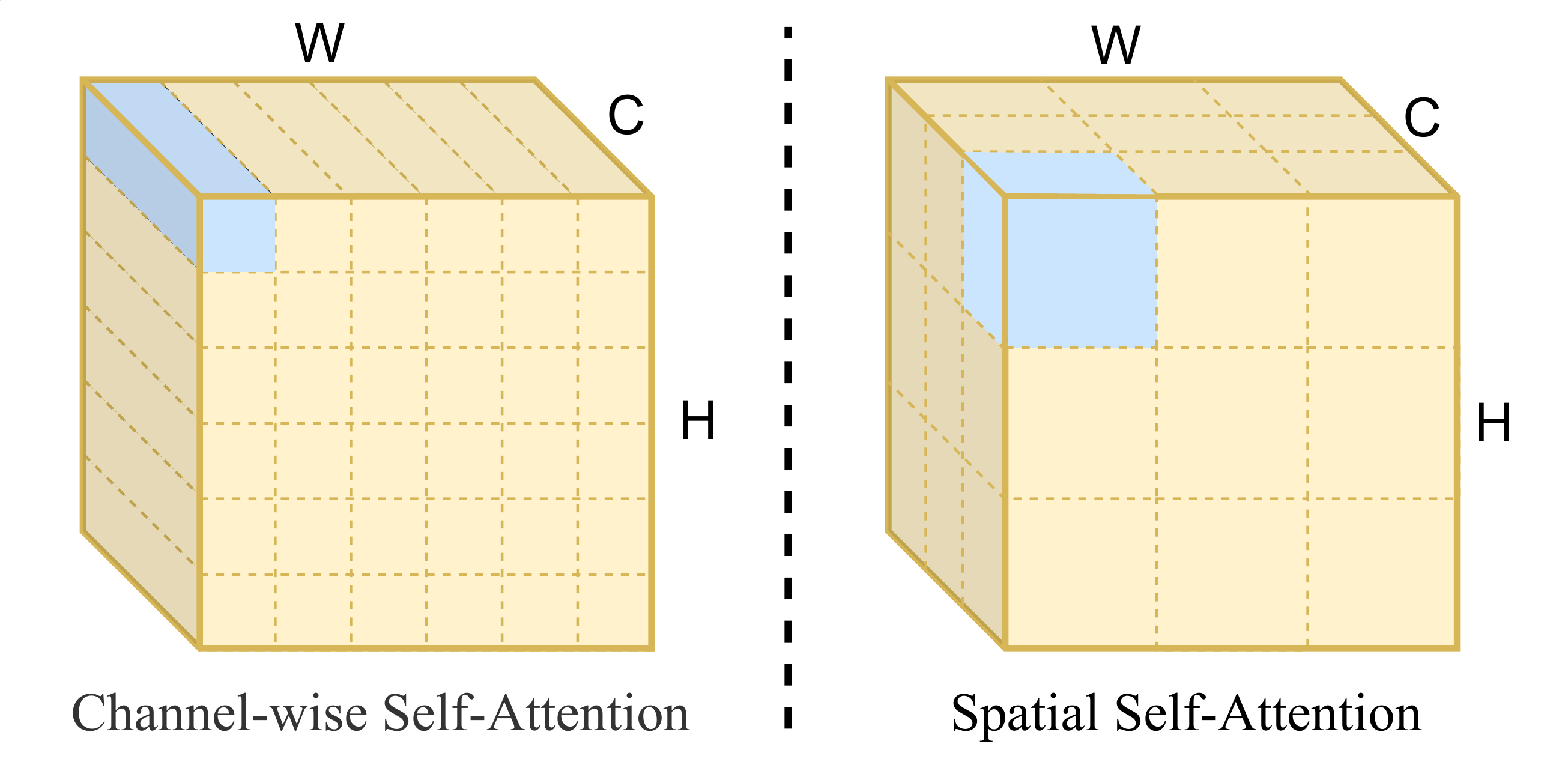}
    \subcaption[]{Existing self-attention}
    \label{fig:intro-b}
    \end{subfigure}
    \end{minipage}
  \begin{minipage}{0.342\linewidth}
    \centering
    \begin{subfigure}{\linewidth}
    \includegraphics[width=1\linewidth]{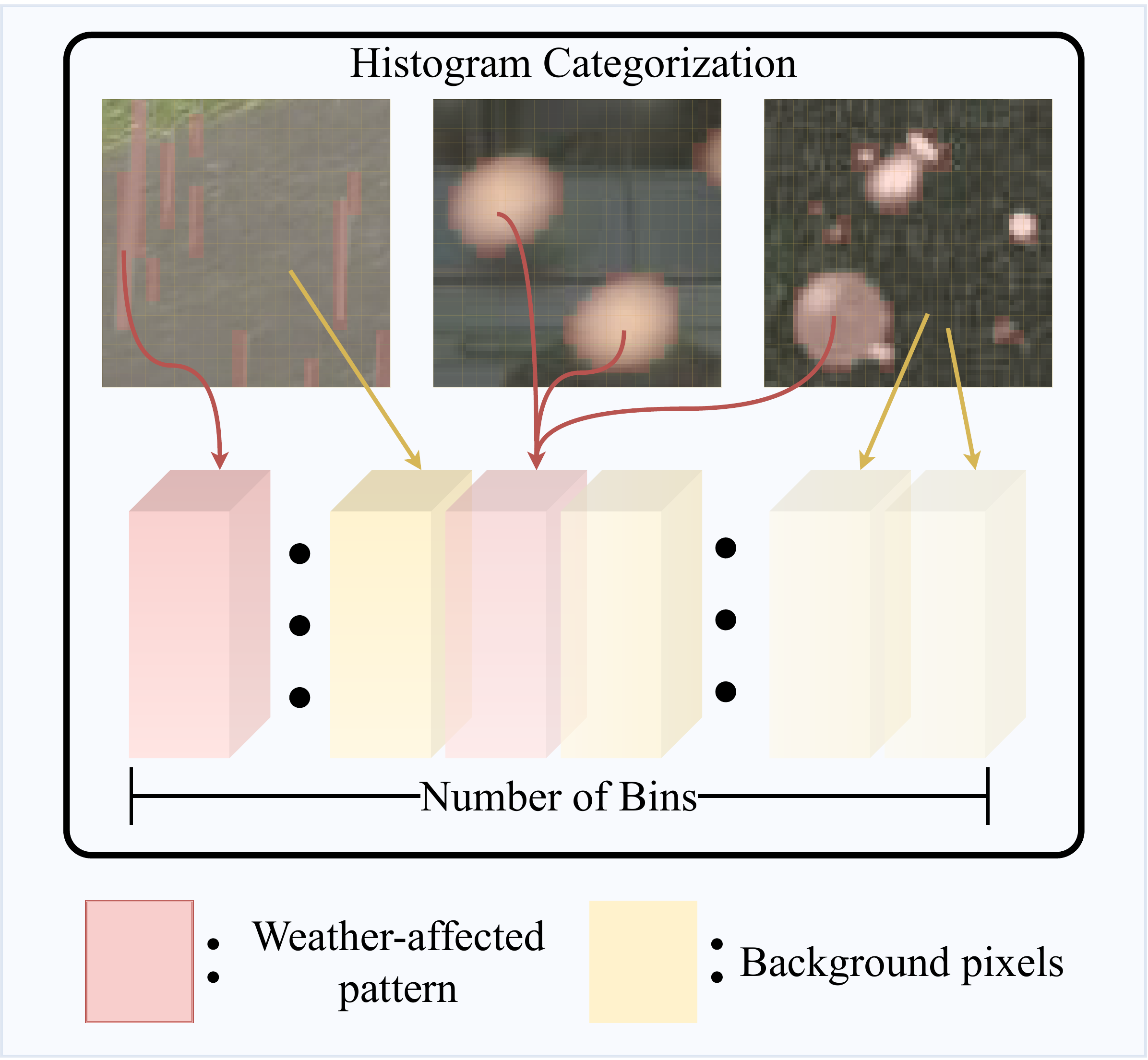}
    \subcaption[]{Histogram self-attention}
    \label{fig:intro-c}
    \end{subfigure}
    \end{minipage}
    \caption{Given weather-degraded images in (a), traditional transformers perform self-attention either along the channel dimension or within a fixed-range block as shown in (b). 
   In contrast, we observe that weather-induced degradation patterns tend to be similar but distinct from the background. 
   So we categorize pixels affected by adverse weather and background pixels into distinct bins based on descending intensities (as depicted in (c)) and then conducts self-attention within and between bins.}
   \label{fig:intro}
\end{figure}

\section{Introduction}
\label{sec:intro}

The field of computer vision witnessed growing interest in restoring images affected by adverse weather conditions like rain, fog, and snow.
These weather conditions significantly degrade visual quality, impacting the performance of downstream tasks such as object detection~\cite{ren2015faster,carion2020end}, and depth estimation~\cite{fu2018deep,godard2017unsupervised}.
The restoration of images under adverse weather is thereby a vital problem for the sake of vision aesthetics and safety.

Early works leverage weather-related priors to model statistical characteristic of degradation and remove adverse weathers~\cite{kang2011automatic,li2016rain,ancuti2013single,berman2016non,zhang2006rain,jiang2017novel,wei2017should,zhu2017joint,you2015adherent}. 
Subsequently, convolutional neural networks (CNNs) have emerged as powerful tools for addressing deraining~\cite{fu2017removing,yang2017deep,liu2018erase,chen2018robust,li2018recurrent,qian2018attentive,wang2019spatial,yasarla2019uncertainty,zhang2018density,zhang2019image}, dehazing~\cite{ren2016single,wu2021contrastive,zhang2018density,zhang2019joint,zhang2021hierarchical,ren2018gated,li2018benchmarking,shao2020domain,li2019semi} and desnowing~\cite{liu2018desnownet,ren2017video,zhang2021deep,li2021online}. 
However, the need of separately training networks for each task and the complexity of switching among multiple models present challenges for real-world applications. 
Li \textit{et al}.~\cite{li2020all} thus introduced the challenge of adverse weather removal, which entails the restoration of images affected by various weather conditions using a single unified model. 

Recently, Transformer-based approaches have also been investigated for the adverse weather removal task, surpassing the efficacy of CNNs~\cite{chen2023learning,valanarasu2022transweather,wang2023gridformer,guo2022image}.
Nonetheless, these Transformer-based methods usually make concessions regarding efficient memory utilization by confining self-attention operations to a fixed spatial range or solely within the channel dimension, as depicted in Figure~\ref{fig:intro-b}.
This compromise impedes the inherent potential of Transformers, which was originally designed for superior global feature modeling, and consequently, it leads to a deterioration in the performance of restoration.

To address these problems, based on the observation that weather-induced degradation often exhibits common patterns shown in Figure~\ref{fig:intro-a}, we develop an efficient \textbf{Histo}gram Trans\textbf{former} for unified adverse weather removal, named Histoformer.
Specifically, we introduce a Dynamic-range Histogram Self-Attention (DHSA) module, which endows self-attention with a dynamic-range spatial receptive field.
We categorize pixel values proximate in intensity yet varied in spatial location into histogram bins.
Self-attention is executed across the dimension of bin or frequency, whose process is illustrated in Figure~\ref{fig:intro-c}. 
To facilitate comprehensive feature extraction on both local and global scales, we devise two ways of reshaping for histogram self-attention: bin-wise histogram reshaping (BHR) and frequency-wise histogram reshaping (FHR).
In BHR, the number of bins is configured to incorporate pixels spanning a more comprehensive intensity range, thereby facilitating global feature integration. 
In FHR, the number of frequencies is assigned such that each bin focuses on limited number of pixels, enhancing the utility of finer features. Consequently, the histogram self-attention attains the capability of modeling spatially dynamic ranges effectively.

To enable the convolution to extract dynamically-located weather-related dependencies, we develop a dynamic-range convolution layer, which involves sequential horizontal and vertical pixel sorting prior to the application of separable convolution.
In order to capture multi-scale and multi-range information embedded within feature matrices, we introduce a Dual-scale Gated Feed-Forward (DGFF) module, enhancing its ability to model the visual characteristics effectively.
Additionally, we note that conventional loss functions primarily focus on pixel-level closeness, overlooking the correlation at overall patch level. 
Consequently, we propose to leverage the Pearson correlation coefficient~\cite{cohen2009pearson} to ensure the reconstruction of the linear relationship between restored and clean images.


Our contribution can be summarized in three folds:
\begin{itemize}
    \item We propose a novel transformer targeted for unified adverse weather removal, equipped with a new histogram self-attention. 
    It possesses dynamic-range spatial attention to weather-induced obstructions and thus can achieve degradation removal globally and efficiently.
    \item To capture multi-range information, we present a dual-scale feed-forward module. 
    To enhance the comprehensive linear association between the recovered and ground-truth images, we develop a correlation loss.
    \item Our method attains state-of-the-art performance across various datasets. Additionally, we substantiate the efficacy of the proposed approach to restore real-world images and bolster the downstream application of detection.
\end{itemize}
\section{Related Work}
Extensive research has been dedicated to addressing adverse weather removal challenges in computer vision, including tasks like deraining~\cite{fu2017removing,yang2017deep,liu2018erase,chen2018robust,li2018recurrent,wang2019spatial,yasarla2019uncertainty,zhang2018density,zhang2019image,sun2023event}, dehazing~\cite{ren2016single,wu2021contrastive,zhang2018density,zhang2019joint,zhang2021hierarchical,ren2018gated,li2018benchmarking,shao2020domain,li2019semi,sun2022rethinking}, desnowing~\cite{liu2018desnownet,ren2017video,zhang2021deep,li2021online}, raindrop removal~\cite{qian2018attentive,quan2019deep,you2015adherent,zhang2021dual} and All-in-One weather removal~\cite{valanarasu2022transweather,ozdenizci2023restoring,li2020all,li2022all}. 

\paragraph{Rain Streak Removal.}
The evolution of approaches is notable in rain streak removal techniques in computer vision. 
Kang \textit{et al}.~\cite{kang2011automatic} pioneered a single image deraining method using bilateral filters to decompose images into low and high-frequency components. 
However, recent advancements have seen a dominance of deep neural networks. 
An early deep CNN was introduced by Fu \textit{et al}.~\cite{fu2017removing} for extracting features from the high-frequency rain component, while Yang \textit{et al}.~\cite{yang2017deep} utilized recurrent networks to decompose rain layers and remove various streak types. 
Li \textit{et al}.~\cite{li2019heavy} proposed a method that addresses rain streaks and veiling effects in heavy rain scenes by integrating physics-based rain models and adversarial learning. 
A conditional generative adversarial network was also employed to solve rain streak removal~\cite{zhang2019image}. 
Yasarla \textit{et al}.~\cite{Yasarla2020syn2real} explored Gaussian processes for transfer learning from synthetic to real-world rain data. 
Quan \textit{et al}.~\cite{quan2021removing} used a cascaded network to remove both rain streaks and raindrops. 
Recently an image deraining Transformer~\cite{xiao2022image} featuring a dual Transformer architecture was intricately formulated, incorporating both window-based and spatial-based mechanisms, thereby attaining exemplary outcomes.
A sparse deraining Transformer is also proposed to enhance feature aggregation~\cite{chen2023learning}.

\paragraph{Raindrop Removal.}
Raindrop removal from single images has been addressed through various methods, with some relying on traditional hand-crafted features. 
An early work incorporated temporal information to address video-based raindrop removal~\cite{you2015adherent}.
Eigen \textit{et al}.~\cite{eigen2013restoring} employed a shallow CNN trained with image pairs containing raindrop-degraded and raindrop-free versions, though the results often exhibited blurriness. 
Qian \textit{et al}.~\cite{qian2018attentive} introduced an attention GAN and a new dataset. 
Their method was later improved by Quan \textit{et al}.~\cite{quan2019deep} via generating attention maps based on mathematical raindrop descriptions and combining them with detected raindrop edges.

\paragraph{Snow Removal.}
Desnow-Net~\cite{liu2018desnownet} was among the pioneering CNN-based approaches for snow removal, followed by Li \textit{et al}.'s stacked dense network~\cite{LI2019stacked} and Chen \textit{et al}.'s JSTASR~\cite{chen2020jstasr}, which introduced a size and transparency aware method. 
More recently, DDMSNet~\cite{zhang2021deep} introduced a dense multi-scale network that leverages semantic and geometric priors to enhance snow removal. 
A hierarchical decomposition paradigm involving the dual-tree wavelet transform for snow removal is also proposed~\cite{chen2021all}.
Chen \textit{et al}.~\cite{chen2022snowformer} designed SnowFormer, a framework that used cross-attentions to establish local-global context interaction.

\paragraph{Fog Removal.}
Li \textit{et al}.~\cite{li2017aod} presented a CNN that takes into account both atmospheric luminosity and transmission maps to conduct dehazing. 
Ren \textit{et al}.~\cite{ren2018gated} advocated a pre-processing approach for hazy image manipulation, thereby engendering multiple input modalities and, in the process, inducting chromatic aberrations as part of their dehazing procedure. 
A hierarchical density-aware network is also introduced, specializing in the domain of image dehazing~\cite{zhang2021hierarchical}.
Zheng \textit{et al}.~\cite{zheng2023curricular} formulated a curriculum-based contrastive regularization dehazing method aimed at fostering agreement within a contrastive space.

\paragraph{All-in-One Weather Removal.}
Some recent works attempted to address various weather-induced degradations by a singular network.
Li \textit{et al}.~\cite{li2020all} proposed an All-in-One network, containing a generator comprising multiple task-specific encoders and a shared decoder. 
Valanarasu \textit{et al}.~\cite{valanarasu2022transweather} presented TransWeather, a transformer-based model featuring a solitary encoder-decoder structure, capable of rejuvenating images afflicted by various atmospheric conditions.
A pipeline for the automatic selection of weather-degraded data was also proposed to enhance existing models~\cite{zhang2023weatherstream}.
Zhu \textit{et al}. developed WGWS-Net~\cite{zhu2023learning_wgwsnet} capable of learning weather-general and weather-specific in two separate stages.
Some other recent works also trial addressing adverse weather removal by adopting probabilistic denoising diffusion model~\cite{ozdenizci2023restoring}, knowledge distillation~\cite{Chen2022MultiWeatherRemoval}, large-scale Pre-trained model~\cite{tan2023exploring}, mixture of experts~\cite{luo2023mowe}, few-shot learning~\cite{kim2023metaweather}, codebooks~\cite{ye2023adverse,wan2023restoring,liu2023learning}, adaptive filters~\cite{park2023all}, knowledge assignment~\cite{wang2023smartassign} and domain translation~\cite{patil2023multi}.

\paragraph{Transformer-based Image Restoration.}
Since the inception of the Vision Transformer (ViT)~\cite{dosovitskiy2020image} for visual recognition, transformers have gained substantial traction across a spectrum of computer vision tasks~\cite{zamir2022restormer,li2024mapping,liu2022video,lyu2023backdoor,Lai_2024_CVPR,liu2021swin,ma2022learning}. 
Particularly within the realm of low-level vision, the Image Processing Transformer~\cite{chen2021pre} exemplifies how pre-training a transformer on extensive datasets can significantly enhance performance for low-level applications. 
U-former~\cite{wang2022uformer}, on the other hand, introduced a transformer architecture based on the U-Net design for restoration tasks. 
Swin-IR~\cite{liang2021swinir} employed the Swin Transformer~\cite{liu2021swin} for image restoration.
Some latest Transformer-based methods were proposed for deraining~\cite{liang2022drt,chen2023learning}, desnowing~\cite{chen2022snowformer},  dehazing~\cite{song2023vision,guo2022image,liu2023data} and All-in-One weather removal~\cite{valanarasu2022transweather,wang2023gridformer}.

Unlike the existing Transfomer-based approaches whose self-attention is applied within either fixed spatial ranges or merely channel dimension, our method enables dynamic-range spatial attention to adaptively focus on weather-induced degradation with similar patterns.

\section{Method}




\subsection{Overall Architecture}

\begin{figure*}[t]
\setlength{\abovecaptionskip}{5pt}
\setlength{\belowcaptionskip}{0pt}
  \centering
   \includegraphics[width=1\linewidth]{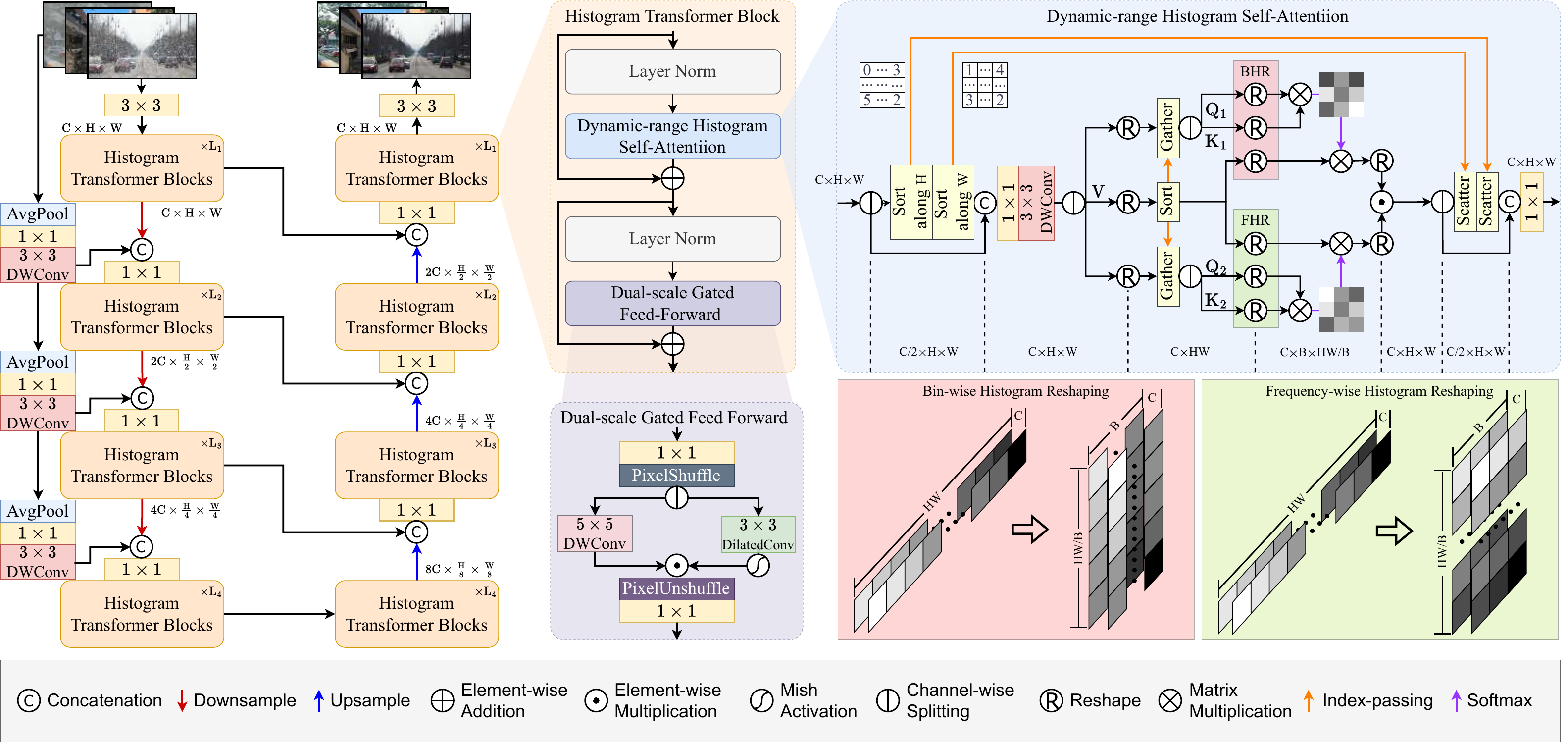}
   \caption{The overall architecture of our Histoformer for weather removal. The main component is the Histogram Transformer block, and it comprises the Dynamic-range Histogram Self-Attention (DHSA) module and the Dual-scale Gated Feed-Forward (DGFF) module. Within DHSA, we present two types of reshaping mechanism, \textit{i.e.,} Bin-wise Histogram Reshaping and Frequency-wise Histogram Reshaping.}
   \label{fig:structure}
\end{figure*}

The overall architectural framework of our Histoformer is illustrated in Figure~\ref{fig:structure}.
Suppose the input is a low-quality image \(I^{lq} \in \mathbb{R}^{3\times H\times W}\),
we pass the input through a \(3\times 3\) convolution to conduct the overlapping image patch-embedding. 
Within both the encoder and decoder of the network backbone, we arrange
Histogram Transformer Blocks (HTBs) to extract intricate features and capture dynamically distributed degradation factors. 
Within the same stage, encoders and decoders are interlinked through skip-connections, 
thereby establishing connections between consecutive intermediate features to enhance the stability of the training process.
Between each stage, we apply pixel-unshuffle and pixel-shuffle operations for the purpose of feature down-sampling and up-sampling.

Within each HTB, we introduce Dynamic-range Histogram Self-Attention (DHSA) to extract spatially dynamic weather degradation and enhance both local and global feature aggregation. 
Moreover, a Dual-scale Gated Feed-Forward (DGFF) module is integrated into the HTB to enrich the representation of multi-range features, contributing to the process of image restoration.
During each stage of encoding phases, our model is equipped with a crude skip-connection for supplementing original features from input, comprised of a sequence of operations, including average pooling, pixel-wise convolution, and depth-wise convolution.
%
We start the crude skip-connection after the first stage, 
and this setup enables the encoders to focus more effectively on learning the weather-induced residuals.
Through this hybrid formulation, Histoformer is empowered to exploit both the adaptive contents of weather-irrelevant background and the inherent characteristics of weather-degraded patterns, facilitating the separation of undesired degradation from the latent clear background.

\subsection{Histogram Transformer Block}
\label{sec:HTB}
As the key component of our Histoformer, HTB contains two pivotal modules, \textit{i.e.,} DHSA and DGFF.
These two components are arranged to interact with layer normalization and can be formulated as follows:
\begin{align}
    F_{l} &= F_{l-1} + {\rm DHSA} \left({\rm LN} \left(F_{l-1}\right)\right),\\
    F_{l} &= F_{l} + {\rm DGFF} \left({\rm LN} \left(F_{l}\right)\right),
\end{align}
where \({\rm LN}\) denotes layer normalization and \(F_l\) represents the feature at \(l\)-th stage.
The details of DHSA and DGFF are presented in Section~\ref{sec:DHSA} and \ref{sec:DGFF} respectively.

\subsubsection{Dynamic-range Histogram Self-Attention}
\label{sec:DHSA}
To better capture dynamically distributed weather-induced degradation, we develop a Dynamic-range Histogram Self-Attention (DHSA) module. 
This module consists of a process involving dynamic-range convolution, which reorders the spatial distribution of fractional features, and a dual-path histogram self-attention mechanism that combines global and local dynamic feature aggregation. 
Prior to the final output projection of a \(1\times 1\) point-wise convolution, the reordered features are sorted back into their original locations to maintain spatial consistency.

\paragraph{Dynamic-range Convolution.}
Traditional convolution operations employ fixed kernel sizes, resulting in a limited receptive field range and consequently perform local and small-range computations.
This restricted scope of convolution, which primarily focuses on local information, does not naturally align with the self-attention mechanism's capacity to model long-range dependencies.
To address this limitation, we devise a dynamic-range convolution technique by meticulously reordering the input features prior to the traditional convolution operation. 
Given an input feature \(F\in \mathbb{R}^{C\times H\times W}\), we divide it into two branches along the channel dimension, namely \(F_{1}\) and \(F_{2}\).
For the first branch of features, we perform sorting operations both horizontally and vertically, subsequently concatenating the sorted features with the second branch of feature.
The resulting recombined features are then passed through the subsequent separable convolution. 
The entire process is articulated as follows:
\begin{equation}
\begin{split}
    &F_{1}, F_{2} = {\rm Split} (F),\ 
    F_{1} = {\rm Sort_{v}} ({\rm Sort_{h}} (F_{1})), \\
    &F = {\rm Conv^d_{3\times 3}}({\rm Conv}_{1\times 1}\left( {\rm Concat}(F_{1}, F_{2}) \right)),
\end{split}
\end{equation}
where \({\rm Conv}_{1\times 1}\) is \(1\times 1\) point-wise convolution, \({\rm Conv^d_{3\times 3}}\) represents \(3\times 3\) depth-wise convolution, \({\rm Concat}\) is the concatenation operation along channel, \({\rm Split}\) denotes the operation of splitting features along channel dimension, and \({\rm Sort}_{i\in \{\rm h, v\}}\) represents the horizontal or vertical sorting operation. 
This approach organizes pixels of high and low intensities into regular patterns at the diagonal corners of the matrices, thereby allowing convolution to perform computations across dynamic ranges.
Given that weather-induced degradation typically exhibits closely related patterns, degraded pixels tend to concentrate in neighboring locations, separated with those clean pixels.
As a result, this arrangement enables convolution kernels to partially focus on preserving clean information and separately recovering degraded features.

\paragraph{Histogram Self-Attention.}
Existing vision Transformers~\cite{xiao2022image,zamir2022restormer,xiao2022image,wang2022uformer,chen2023learning,zhao2023comprehensive} typically leverage fixed range of attention or merely the attention along channel dimension due to the compromise of computation and memory efficiency. 
However, the fixed setting restricts the self-attention to span adaptively long range to associate desired features.
We notice that weather-induced degradation causes similar patterns and that those pixels containing either background feature or weather degradation of different intensities had better be assigned with various extents of attention. 
We thus propose a histogram self-attention mechanism to categorize spatial elements into bins and allocate varying attention within and across bins. 
For the sake of parallel computing, we set each bin contains identical number of pixels during implement.

Given the output of dynamic-range convolution, we can separate them into Value feature \(V\in \mathbb{R}^{C\times H\times W}\) and two pairs of Query-key \(F_{QK,1},F_{QK,2}\in \mathbb{R}^{2C\times H\times W}\) subsequently passing to two branches. 
We firstly sort the sequence of \(V\) and based on its index arrange the Query-Key pairs accordingly, expressed as follows:
\begin{equation}
\begin{split}
    &V, d = {\rm Sort}\left({\rm \mathbf{R}_{C\times H\times W}^{C\times HW}} (V)\right), \\
    &Q_1, K_1 = {\rm Split}\left({\rm Gather}\left({\rm \mathbf{R}_{C\times H\times W}^{C\times HW}} (F_{QK,1}), d\right) \right), \\
    &Q_2, K_2 = {\rm Split}\left({\rm Gather}\left({\rm \mathbf{R}_{C\times H\times W}^{C\times HW}} (F_{QK,2}), d\right) \right),
\end{split}
\end{equation}
where \({\rm \mathbf{R}_{C\times, H\times W}^{C\times HW}}\) represents the operation of reshaping features from \(\mathbb{R}^{C\times H\times W}\) to \(\mathbb{R}^{C\times HW}\), \(d\) is the index of sorted Value, and \({\rm Gather}\) denotes the operation of retrieving elements of tensor based on a given index.

Then given the number of bins \(B\), we reshape the sorted features from \(C\times HW\) into \(C\times B\times HW/B\). 
To extract both global and local information, we define two types of reshaping, i.e., bin-wise histogram reshaping (BHR) and frequency-wise histogram reshaping (FHR). 
The first is to assign the number of bins equal to \(B\) and each bin contains \(HW/B\) elements, while the second is to set the frequency of each bin equal to \(B\) and the number of bins is \(HW/B\).
By this way, we can extract large-scale information by BHR where each bin contains large number of dynamically located pixels and fine-grained information by FHR where each bins contains modicum pixels neighboring in terms of intensity.
The two pairs of Query-Key features are passed through two types of reshaping and subsequent self-attention process respectively, and their outputs are element-wisely multiplied to yield the final output. 
The process can be formulated as the following expressions:
%
\begin{equation}
    \begin{split}
        &{\rm A_B} = {\rm softmax}\left(\frac{{\rm \mathbf{R}_B}(Q_1){\rm \mathbf{R}_B}(K_1)^{\top}} {\sqrt{k}}\right) {\rm \mathbf{R}_B}(V), \\
        &{\rm A_F} = {\rm softmax}\left(\frac{{\rm \mathbf{R}_F}(Q_2){\rm \mathbf{R}_F}(K_2)^\top}{\sqrt{k}}\right) {\rm \mathbf{R}_F}(V), \\
        &{\rm A} = {\rm A_B}  \odot {\rm A_F}, 
    \end{split}
\end{equation}
where \(k\) is the number of heads, \({\rm \mathbf{R}_{\textit{i}\in \{B,F\}}}\) denotes the reshaping operation of either BHR or FHR, and \({\rm A_{\textit{i}\in\{B,F\}}}\) represents the obtained attention map.

\subsubsection{Dual-scale Gated Feed-Forward}
\label{sec:DGFF}

Previous studies~\cite{zamir2022restormer,xiao2022image,wang2022uformer,chen2023learning} typically leverage single-range or single-scale convolution in the standard feed-forward network to bolster local context. 
Nonetheless, these methods often disregard the correlations among dynamically distributed weather-induced degradation. 
In practice, multi-scale information can be extracted by not only enlarging the kernel size but also leveraging the dilation mechanism~\cite{yu2016multiscale,yu2017dilated,li2023dilated}. 
As a result, we conceive a Dual-scale Gated Feed-Forward (DGFF) module, which integrates two distinct multi-range and multi-scale depth-wise convolution pathways within the transmission process.

Given an input tensor \(F_{l} \in \mathbb{R}^{C\times H\times W}\), we initially employ a point-wise convolution operation to augment the channel dimension by a factor of \(r\). 
Following this augmentation, the expanded tensor is directed into two parallel branches. 
Throughout the feature transformation process, \(5\times 5\) and dilated \(3\times 3\) depth-wise convolutions are employed to enhance the extraction of multi-range and multi-scale information.
Following the gating mechanism~\cite{dauphin2017language}, the output of the second branch after passing through an activation act as a gating map for the other branch.
Thus, the complete feature fusion process within the DGFF module is formulated as follows:
\begin{equation}
\begin{split}
    &F_{l,1}, F_{l,2} = {\rm Split}\left( {\rm Shuffle} ({\rm Conv}_{1\times 1} (F_{l}))\right), \\
    &F_{l,1} = {\rm Conv^d_{5\times 5}} (F_{l,1}),\ 
    F_{l,2} = {\rm Conv^{d,dilated}_{3\times 3}} (F_{l,2}), \\
    &F_{l+1} = {\rm Conv}_{1\times 1}\left({\rm Unshuffle} \left({\rm Mish}(F_{l,2}) \odot F_{l,1} \right)\right),
\end{split}
\end{equation}
where \({\rm Conv^{d}_{5\times 5}}\) represents \(5\times 5\) depth-wise convolution, \({\rm Conv^{d,dilated}_{3\times 3}}\) is \(3\times 3\) dilated depth-wise convolution, \({\rm Shuffle}\) and \({\rm Unshuffle}\) represent respectively the operations of pixel-shuffling and unshuffling, \({\rm Mish}\) denotes the Mish activation~\cite{misra2019mish}, and \(F_{l+1}\) is the output of current stage passing to \(l+1\)-th stage.

\subsection{Reconstruction Loss and Correlation Loss}

We use the \(L_1\) norm of the pixel-wise difference between the restored high-quality image \(I^{hq}\) and ground-truth \(I^{gt}\) as the reconstruction loss, i.e.,
\begin{equation}
    \mathcal{L}_{rec} = \left\|I^{hq} - I^{gt}\right\|_1.
\end{equation}
\begin{table*}[t]
\setlength{\abovecaptionskip}{5pt}
\setlength{\belowcaptionskip}{0pt}
\tiny
\tabcolsep=0.0mm
    \centering
    \caption{Quantitative comparisons on three weather removal tasks in terms of PSNR and SSIM, where higher values indicate better performance. 
    The top halves of tables display the results of task-specific methods, while the bottom halves present evaluations of the unified multi-weather models. 
    The best and the second best results are in \textbf{bold} and \underline{underlined}. Those with \(^*\) indicate the methods whose source codes are unavailable.}
    \label{tab:result}
    \scalebox{0.9}{
    \begin{subtable}[t]{0.462\linewidth}
        \caption{Image Desnowing}\label{tab:snow}%
        \begin{tabular}{lcccc}
            \toprule
                  & \multicolumn{2}{c}{Snow100K-S~\cite{liu2018desnownet}} & \multicolumn{2}{c}{Snow100K-L~\cite{liu2018desnownet}} \\
                  & PSNR  & SSIM  & PSNR  & SSIM \\
            \midrule
            SPANet~\cite{wang2019spatial} & 29.92 & 0.8260 & 23.70 & 0.7930 \\
            JSTASR~\cite{chen2020jstasr} & 31.40 & 0.9012 & 25.32 & 0.8076 \\
            RESCAN~\cite{li2018recurrent} & 31.51 & 0.9032 & 26.08 & 0.8108 \\
            DesnowNet~\cite{liu2018desnownet} & 32.33 & 0.9500 & 27.17 & 0.8983 \\
            DDMSNet~\cite{zhang2021deep} & 34.34 & 0.9445 & 28.85 & 0.8772 \\
            NAFNet~\cite{chen2022simple} & 34.79 & 0.9497 & 30.06 & 0.9017 \\
            Restormer~\cite{zamir2022restormer} & 36.01 & 0.9579 & 30.36 & 0.9068 \\
            \midrule
            All-in-One~\cite{li2020all}\(^*\) & -     & -     & 28.33 & 0.8820 \\
            TransWeather~\cite{valanarasu2022transweather} & 32.51 & 0.9341 & 29.31 & 0.8879 \\
            Chen \textit{et al.}~\cite{Chen2022MultiWeatherRemoval} & 34.42 & 0.9469 & 30.22 & 0.9071 \\
            WGWSNet~\cite{zhu2023learning_wgwsnet} & 34.31 & 0.9460 & 30.16 & 0.9007 \\
            WeatherDiff\(_{64}\)~\cite{ozdenizci2023restoring}  & 35.83 & 0.9566 & 30.09 & 0.9041 \\
            WeatherDiff\(_{128}\)~\cite{ozdenizci2023restoring}  & 35.02 & 0.9516 & 29.58 & 0.8941 \\
            AWRCP~\cite{ye2023adverse}\(^*\) & \underline{36.92} & \underline{0.9652} & \underline{31.92} & \textbf{0.9341} \\
            Histoformer (Ours) & \textbf{37.41} & \textbf{0.9656} & \textbf{32.16} & \underline{0.9261} \\
            \bottomrule
        \end{tabular}%
    \end{subtable}
    \hfill
    \begin{subtable}[t]{0.328\linewidth}
        \caption{Deraining \& Dehazing}\label{tab:rainfog}%
        \begin{tabular}{lcc}
            \toprule
                  & \multicolumn{2}{c}{Outdoor-Rain~\cite{li2019heavy}} \\
                  & \multicolumn{1}{c}{PSNR} & \multicolumn{1}{c}{SSIM} \\
            \midrule
            CycleGAN~\cite{zhu2017cyclegan} & 17.62 & 0.6560 \\
            pix2pix~\cite{isola2017pix2pix} & 19.09 & 0.7100 \\
            HRGAN~\cite{li2019heavy} & 21.56 & 0.8550 \\
            PCNet~\cite{jiang2021pcnet} & 26.19 & 0.9015 \\
            MPRNet~\cite{Zamir2021mprnet} & 28.03 & 0.9192 \\
            NAFNet~\cite{chen2022simple} & 29.59 & 0.9027 \\
            Restormer~\cite{zamir2022restormer} & 30.03 & 0.9215 \\
            \midrule
            All-in-One~\cite{li2020all}\(^*\) & 24.71 & 0.8980 \\
            TransWeather~\cite{valanarasu2022transweather} & 28.83 & 0.9000 \\
            Chen \textit{et al.}~\cite{Chen2022MultiWeatherRemoval} & 29.27  & 0.9147  \\
            WGWSNet~\cite{zhu2023learning_wgwsnet} & 29.32 & 0.9207 \\
            WeatherDiff\(_{64}\)~\cite{ozdenizci2023restoring} & 29.64 & 0.9312 \\
            WeatherDiff\(_{128}\)~\cite{ozdenizci2023restoring} & 29.72 & 0.9216 \\
            AWRCP~\cite{ye2023adverse}\(^*\) & \underline{31.39} & \underline{0.9329} \\
            Histoformer (Ours) & \textbf{32.08} & \textbf{0.9389} \\
            \bottomrule
        \end{tabular}%
    \end{subtable}
    \hfill
    \begin{subtable}[t]{0.320\linewidth}
        \caption{Raindrop Removal}\label{tab:raindrop}%
        \begin{tabular}{lrr}
            \toprule
                  & \multicolumn{2}{c}{RainDrop~\cite{qian2018attentive}} \\
                  & \multicolumn{1}{c}{PSNR} & \multicolumn{1}{c}{SSIM} \\
            \midrule
            pix2pix~\cite{isola2017pix2pix} & 28.02 & 0.8547 \\
            DuRN~\cite{liu2019durn}  & 31.24 & 0.9259 \\
            RaindropAttn~\cite{quan2019deep} & 31.44 & 0.9263 \\
            AttentiveGAN~\cite{qian2018attentive} & 31.59 & 0.9170 \\
            IDT~\cite{xiao2022image}   & 31.87 & 0.9313 \\
            MAXIM~\cite{tu2022maxim} & 31.87 & 0.9352 \\
            Restormer~\cite{zamir2022restormer} & 32.18 &   \underline{0.9408} \\
            \midrule
            All-in-One~\cite{li2020all}\(^*\) & 31.12 & 0.9268 \\
            TransWeather~\cite{valanarasu2022transweather} & 30.17 & 0.9157 \\
            Chen \textit{et al.}~\cite{Chen2022MultiWeatherRemoval} & 31.81 & 0.9309 \\
            WGWSNet~\cite{zhu2023learning_wgwsnet} & \underline{32.38} & 0.9378 \\
            WeatherDiff\(_{64}\)~\cite{ozdenizci2023restoring} & 30.71 & 0.9312 \\
            WeatherDiff\(_{128}\)~\cite{ozdenizci2023restoring} & 29.66 & 0.9225 \\
            AWRCP~\cite{ye2023adverse}\(^*\) & 31.93 & 0.9314 \\
            Histoformer (Ours) & \textbf{33.06} & \textbf{0.9441} \\
            \bottomrule
        \end{tabular}%
    \end{subtable}
    }
\end{table*}

Furthermore, we notice that the \(\mathcal{L}_{rec}\) only regulates the pixel-level similarity between the restored image and the ground-truth, while neglecting the patch-level linear correlations. 
The innate relationships of intensity within the image are disrupted by the consistent patterns of weather-induced degradation. 
By emulating the intensity relationships within the ground-truth, we compel the degraded pixels to occupy their original positions according to the original intensity ranking.
Consequently, we introduce the Pearson correlation~\cite{cohen2009pearson} between images as a means to regulate the linear relationship, expressed as follows:
\begin{equation}
    \rho\left( I^{hq}, I^{gt}\right) = \frac{ \sum_{i=1}^{3HW} \left(I^{hq}_{i}- \overline{I}^{hq}\right) \left(I^{gt}_{i} - \overline{I}^{gt}\right)}{ 3HW\sigma\left(I^{hq}\right) \sigma\left(I^{gt}\right) },
\end{equation}
where \(I^{\{\cdot\}}_{i}\) represents the \(i\)-th pixel of image, \(\overline{I}^{\{\cdot\}}\) and \(\sigma\left(I^{\{\cdot\}}\right)\)
denotes respectively the mean and the standard deviation of image sequence. 
Its value falls within the \([-1,1]\) range. When two images exhibit perfect correlation, the value of function \(\rho\) attains a value of \(1\), while in the case of negative correlation, its value reaches \(-1\). Hence, we formulate the correlation loss as follows:
\begin{equation}
    \mathcal{L}_{cor} = \frac{1}{2}\left(1-\rho\left( I^{hq}, I^{gt}\right)\right),
\end{equation}
such that \(\mathcal{L}_{cor}=0\) when the recovered image perfectly aligns with the ground-truth. The overall loss function  is thus defined as: 
\begin{equation}
    \mathcal{L} = \mathcal{L}_{rec} +\alpha \mathcal{L}_{cor},
\end{equation}
where \(\alpha\) is the weight of correlation loss.

\begin{figure*}[t]
\setlength{\abovecaptionskip}{5pt}
\setlength{\belowcaptionskip}{0pt}
  \centering
  \begin{minipage}{0.1389\linewidth}
    \centering
    \begin{subfigure}{1\linewidth}
    \includegraphics[width=1\linewidth]{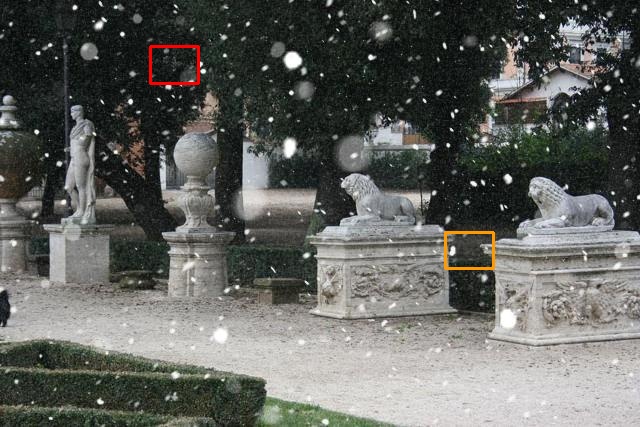}
    \end{subfigure}
    \hspace{-1mm}
     \begin{subfigure}{1\linewidth}
    \includegraphics[width=0.486\linewidth]{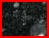}
    \hspace{-1mm}
     \includegraphics[width=0.486\linewidth]{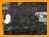}
    \end{subfigure}
    \subcaption[]{Input}
    \label{fig:snow-a}
    \end{minipage}
    \hspace{-1.5mm}
  \begin{minipage}{0.1389\linewidth}
    \centering
    \begin{subfigure}{1\linewidth}
    \includegraphics[width=1\linewidth]{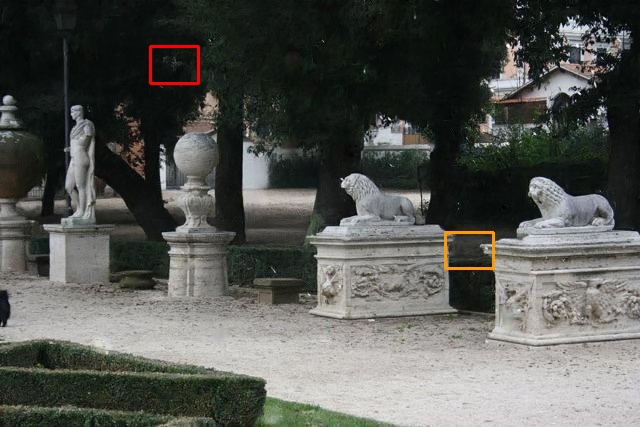}
    \end{subfigure}
    \hspace{-1mm}
     \begin{subfigure}{1\linewidth}
    \includegraphics[width=0.486\linewidth]{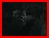}
    \hspace{-1mm}
     \includegraphics[width=0.486\linewidth]{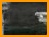}
    \end{subfigure}
    \subcaption[]{\cite{zamir2022restormer}}
    \label{fig:snow-b}
    \end{minipage}
    \hspace{-1.5mm}
  \begin{minipage}{0.1389\linewidth}
    \centering
    \begin{subfigure}{1\linewidth}
    \includegraphics[width=1\linewidth]{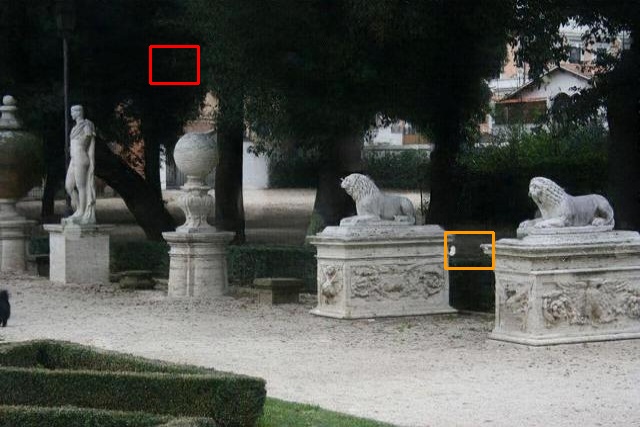}
    \end{subfigure}
    \hspace{-1mm}
     \begin{subfigure}{1\linewidth}
    \includegraphics[width=0.486\linewidth]{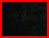}
    \hspace{-1mm}
     \includegraphics[width=0.486\linewidth]{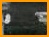}
    \end{subfigure}
    \subcaption[]{\cite{valanarasu2022transweather}}
    \label{fig:snow-c}
    \end{minipage}
    \hspace{-1.5mm}
  \begin{minipage}{0.1389\linewidth}
    \centering
    \begin{subfigure}{1\linewidth}
    \includegraphics[width=1\linewidth]{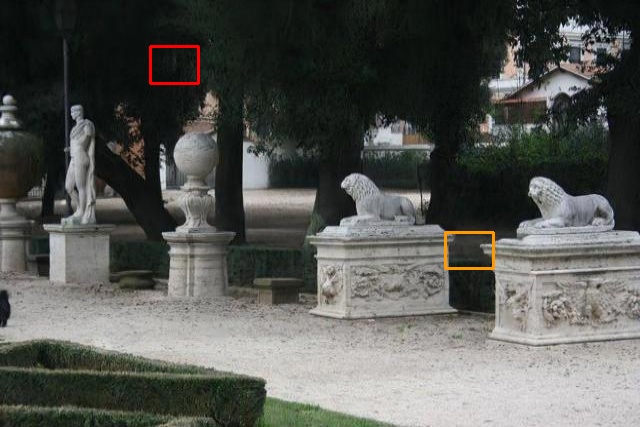}
    \end{subfigure}
    \hspace{-1mm}
     \begin{subfigure}{1\linewidth}
    \includegraphics[width=0.486\linewidth]{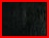}
    \hspace{-1mm}
     \includegraphics[width=0.486\linewidth]{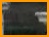}
    \end{subfigure}
    \subcaption[]{\cite{zhu2023learning_wgwsnet}}
    \label{fig:snow-d}
    \end{minipage}
    \hspace{-1.5mm}
  \begin{minipage}{0.1389\linewidth}
    \centering
    \begin{subfigure}{1\linewidth}
    \includegraphics[width=1\linewidth]{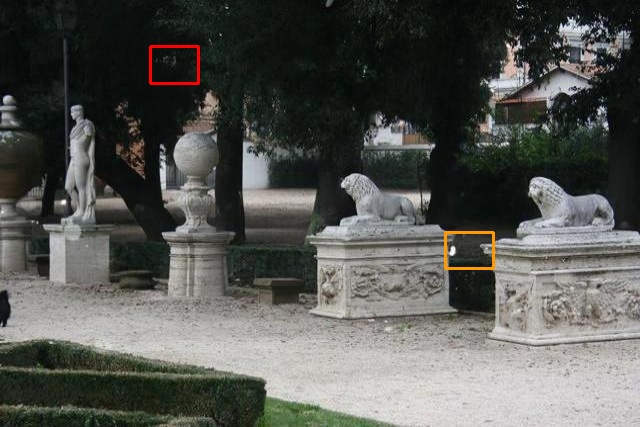}
    \end{subfigure}
    \hspace{-1mm}
     \begin{subfigure}{1\linewidth}
    \includegraphics[width=0.486\linewidth]{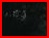}
    \hspace{-1mm}
     \includegraphics[width=0.486\linewidth]{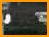}
    \end{subfigure}
    \subcaption[]{\cite{ozdenizci2023restoring}}
    \label{fig:snow-e}
    \end{minipage}
    \hspace{-1.5mm}
  \begin{minipage}{0.1389\linewidth}
    \centering
    \begin{subfigure}{1\linewidth}
    \includegraphics[width=1\linewidth]{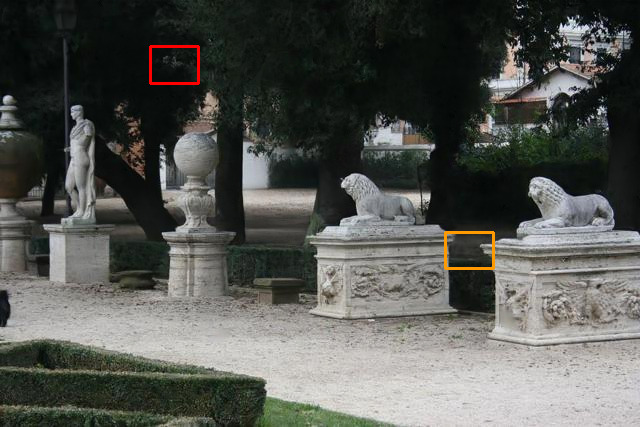}
    \end{subfigure}
    \hspace{-1mm}
     \begin{subfigure}{1\linewidth}
    \includegraphics[width=0.486\linewidth]{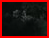}
    \hspace{-1mm}
     \includegraphics[width=0.486\linewidth]{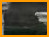}
    \end{subfigure}
    \subcaption[]{Ours}
    \label{fig:snow-f}
    \end{minipage}
    \hspace{-1.5mm}
  \begin{minipage}{0.1389\linewidth}
    \centering
    \begin{subfigure}{1\linewidth}
    \includegraphics[width=1\linewidth]{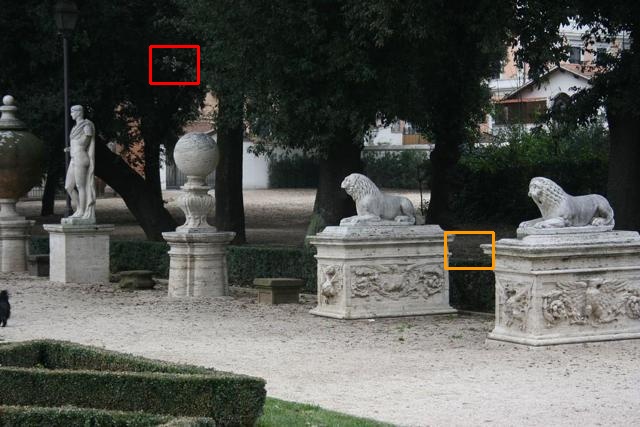}
    \end{subfigure}
    \hspace{-1mm}
     \begin{subfigure}{1\linewidth}
    \includegraphics[width=0.486\linewidth]{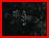}
    \hspace{-1mm}
     \includegraphics[width=0.486\linewidth]{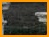}
    \end{subfigure}
    \subcaption[]{Clean}
    \label{fig:snow-g}
    \end{minipage}
  \caption{Visual comparison for desnowing on Snow100K~\cite{liu2018desnownet}. The samples from (b) to (e) are Restormer~\cite{zamir2022restormer}, TransWeather~\cite{valanarasu2022transweather}, WGWSNet~\cite{zhu2023learning_wgwsnet}, WeatherDiff~\cite{ozdenizci2023restoring}.}
  \label{fig:snow}
\end{figure*}
\begin{figure*}[t]
\setlength{\abovecaptionskip}{5pt}
\setlength{\belowcaptionskip}{0pt}
  \centering
  \begin{minipage}{0.1389\linewidth}
    \centering
    \begin{subfigure}{1\linewidth}
    \includegraphics[width=1\linewidth]{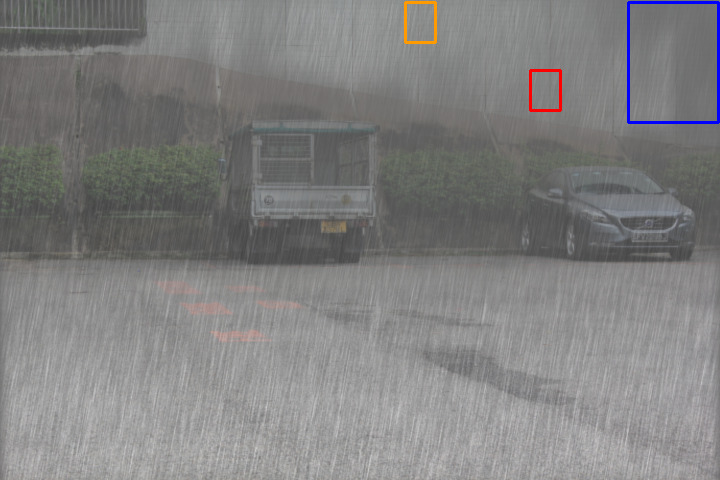}
    \end{subfigure}
    \hspace{-1mm}
     \begin{subfigure}{1\linewidth}
    \includegraphics[width=0.311\linewidth]{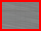}
    \hspace{-1mm}
    \includegraphics[width=0.311\linewidth]{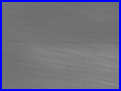}
    \hspace{-1mm}
     \includegraphics[width=0.311\linewidth]{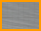}
    \end{subfigure}
    \subcaption[]{Input}
    \label{fig:rainfog-a}
    \end{minipage}
    \hspace{-1.5mm}
  \begin{minipage}{0.1389\linewidth}
    \centering
    \begin{subfigure}{1\linewidth}
    \includegraphics[width=1\linewidth]{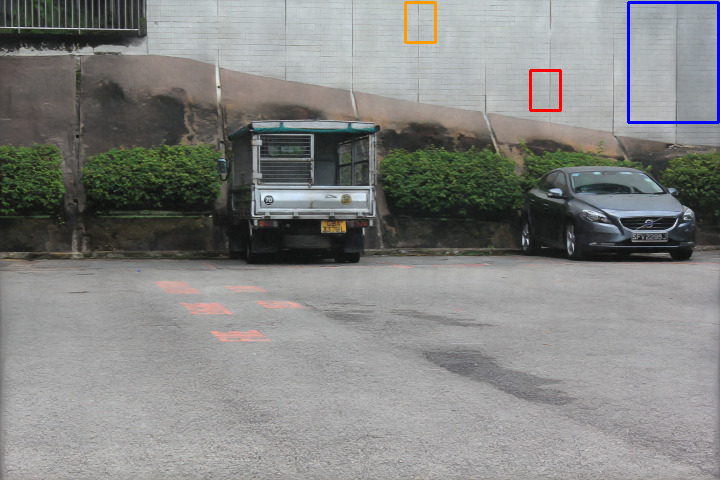}
    \end{subfigure}
    \hspace{-1mm}
     \begin{subfigure}{1\linewidth}
    \includegraphics[width=0.311\linewidth]{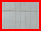}
    \hspace{-1mm}
    \includegraphics[width=0.311\linewidth]{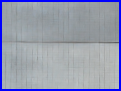}
    \hspace{-1mm}
     \includegraphics[width=0.311\linewidth]{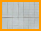}
    \end{subfigure}
    \subcaption[]{\cite{zamir2022restormer}}
    \label{fig:rainfog-b}
    \end{minipage}
    \hspace{-1.5mm}
  \begin{minipage}{0.1389\linewidth}
    \centering
    \begin{subfigure}{1\linewidth}
    \includegraphics[width=1\linewidth]{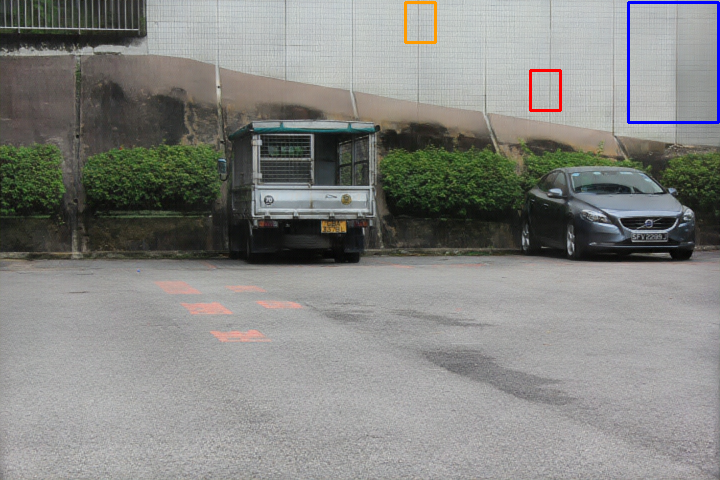}
    \end{subfigure}
    \hspace{-1mm}
     \begin{subfigure}{1\linewidth}
    \includegraphics[width=0.311\linewidth]{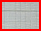}
    \hspace{-1mm}
    \includegraphics[width=0.311\linewidth]{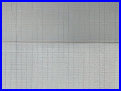}
    \hspace{-1mm}
     \includegraphics[width=0.311\linewidth]{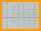}
    \end{subfigure}
    \subcaption[]{\cite{valanarasu2022transweather}}
    \label{fig:rainfog-c}
    \end{minipage}
    \hspace{-1.5mm}
  \begin{minipage}{0.1389\linewidth}
    \centering
    \begin{subfigure}{1\linewidth}
    \includegraphics[width=1\linewidth]{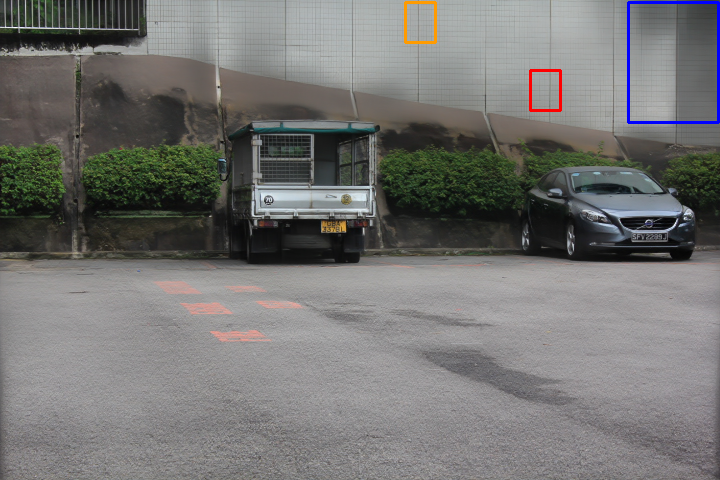}
    \end{subfigure}
    \hspace{-1mm}
     \begin{subfigure}{1\linewidth}
    \includegraphics[width=0.311\linewidth]{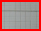}
    \hspace{-1mm}
    \includegraphics[width=0.311\linewidth]{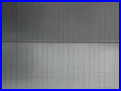}
    \hspace{-1mm}
     \includegraphics[width=0.311\linewidth]{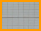}
    \end{subfigure}
    \subcaption[]{\cite{zhu2023learning_wgwsnet}}
    \label{fig:rainfog-d}
    \end{minipage}
    \hspace{-1.5mm}
  \begin{minipage}{0.1389\linewidth}
    \centering
    \begin{subfigure}{1\linewidth}
    \includegraphics[width=1\linewidth]{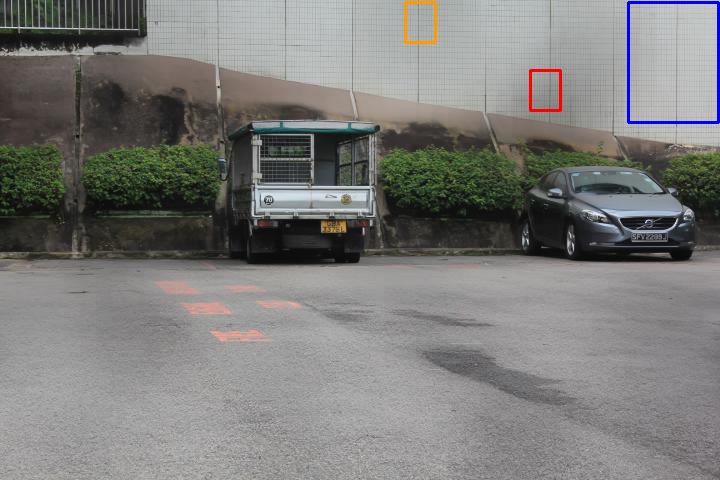}
    \end{subfigure}
    \hspace{-1mm}
     \begin{subfigure}{1\linewidth}
    \includegraphics[width=0.311\linewidth]{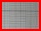}
    \hspace{-1mm}
    \includegraphics[width=0.311\linewidth]{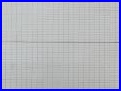}
    \hspace{-1mm}
     \includegraphics[width=0.311\linewidth]{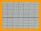}
    \end{subfigure}
    \subcaption[]{\cite{ozdenizci2023restoring}}
    \label{fig:rainfog-e}
    \end{minipage}
    \hspace{-1.5mm}
  \begin{minipage}{0.1389\linewidth}
    \centering
    \begin{subfigure}{1\linewidth}
    \includegraphics[width=1\linewidth]{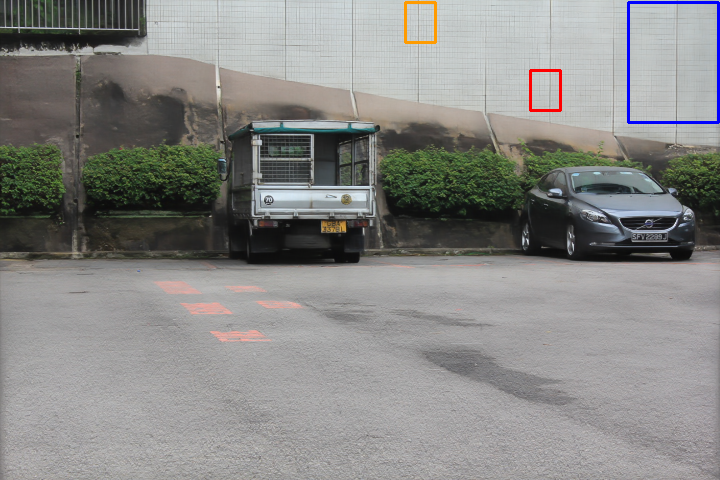}
    \end{subfigure}
    \hspace{-1mm}
     \begin{subfigure}{1\linewidth}
    \includegraphics[width=0.311\linewidth]{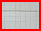}
    \hspace{-1mm}
    \includegraphics[width=0.311\linewidth]{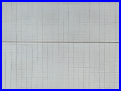}
    \hspace{-1mm}
     \includegraphics[width=0.311\linewidth]{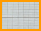}
    \end{subfigure}
    \subcaption[]{Ours}
    \label{fig:rainfog-f}
    \end{minipage}
    \hspace{-1.5mm}
  \begin{minipage}{0.1389\linewidth}
    \centering
    \begin{subfigure}{1\linewidth}
    \includegraphics[width=1\linewidth]{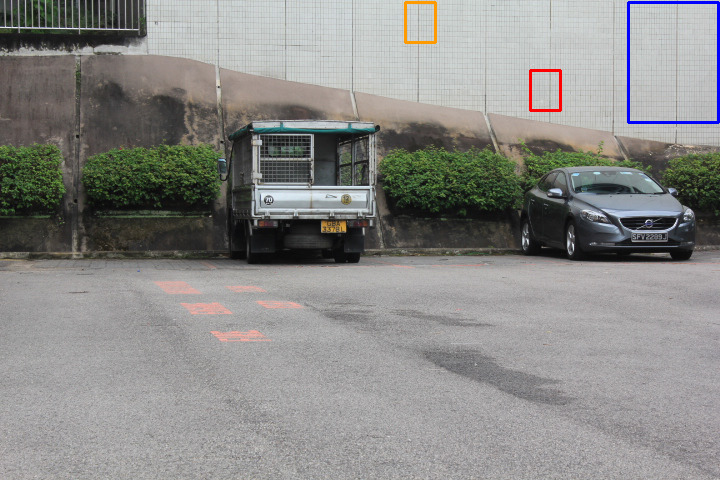}
    \end{subfigure}
    \hspace{-1mm}
     \begin{subfigure}{1\linewidth}
    \includegraphics[width=0.311\linewidth]{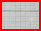}
    \hspace{-1mm}
    \includegraphics[width=0.311\linewidth]{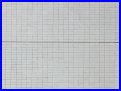}
    \hspace{-1mm}
     \includegraphics[width=0.311\linewidth]{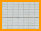}
    \end{subfigure}
    \subcaption[]{Clean}
    \label{fig:rainfog-g}
    \end{minipage}
  \caption{Visual comparison for deraining and dehazing on Outdoor-Rain~\cite{li2019heavy}. The samples from (b) to (e) are Restormer~\cite{zamir2022restormer}, TransWeather~\cite{valanarasu2022transweather}, WGWSNet~\cite{zhu2023learning_wgwsnet}, WeatherDiff~\cite{ozdenizci2023restoring}.}
  \label{fig:rainfog}
\end{figure*}
\begin{figure*}[t]
\setlength{\abovecaptionskip}{5pt}
\setlength{\belowcaptionskip}{0pt}
  \centering
  \begin{minipage}{0.1389\linewidth}
    \centering
    \begin{subfigure}{1\linewidth}
    \includegraphics[width=1\linewidth]{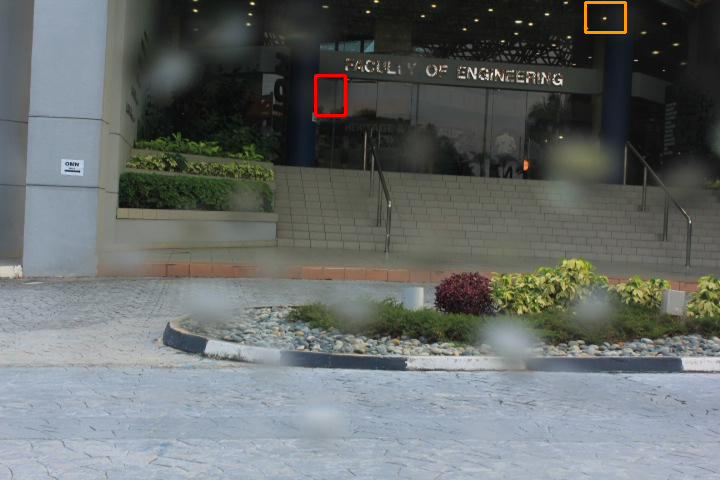}
    \end{subfigure}
    \hspace{-1mm}
     \begin{subfigure}{1\linewidth}
    \includegraphics[width=0.486\linewidth]{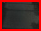}
    \hspace{-1mm}
     \includegraphics[width=0.486\linewidth]{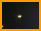}
    \end{subfigure}
    \subcaption[]{Input}
    \label{fig:raindrop-2-a}
    \end{minipage}
    \hspace{-1.5mm}
  \begin{minipage}{0.1389\linewidth}
    \centering
    \begin{subfigure}{1\linewidth}
    \includegraphics[width=1\linewidth]{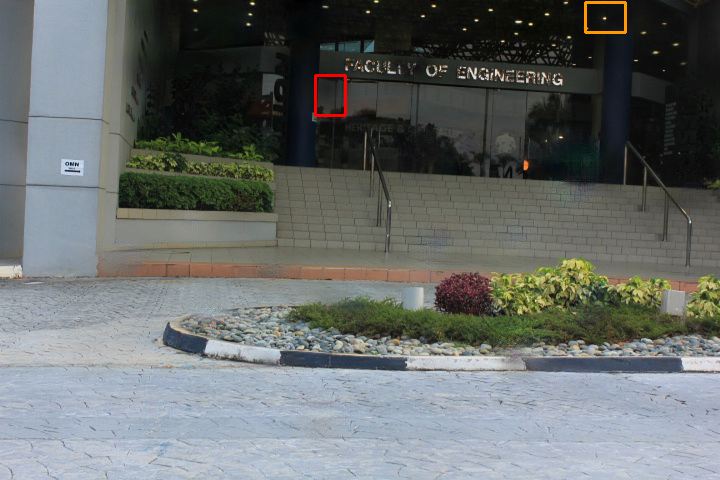}
    \end{subfigure}
    \hspace{-1mm}
     \begin{subfigure}{1\linewidth}
    \includegraphics[width=0.486\linewidth]{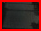}
    \hspace{-1mm}
     \includegraphics[width=0.486\linewidth]{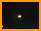}
    \end{subfigure}
    \subcaption[]{\cite{Chen2022MultiWeatherRemoval}}
    \label{fig:raindrop-2-b}
    \end{minipage}
    \hspace{-1.5mm}
  \begin{minipage}{0.1389\linewidth}
    \centering
    \begin{subfigure}{1\linewidth}
    \includegraphics[width=1\linewidth]{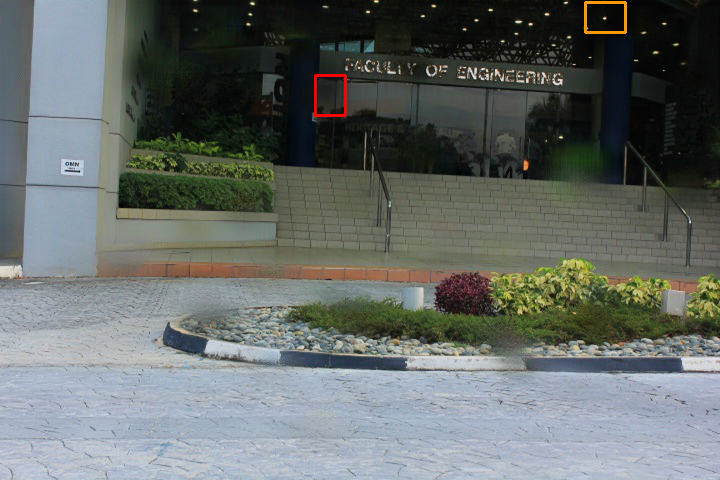}
    \end{subfigure}
    \hspace{-1mm}
     \begin{subfigure}{1\linewidth}
    \includegraphics[width=0.486\linewidth]{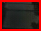}
    \hspace{-1mm}
     \includegraphics[width=0.486\linewidth]{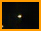}
    \end{subfigure}
    \subcaption[]{\cite{valanarasu2022transweather}}
    \label{fig:raindrop-2-c}
    \end{minipage}
    \hspace{-1.5mm}
  \begin{minipage}{0.1389\linewidth}
    \centering
    \begin{subfigure}{1\linewidth}
    \includegraphics[width=1\linewidth]{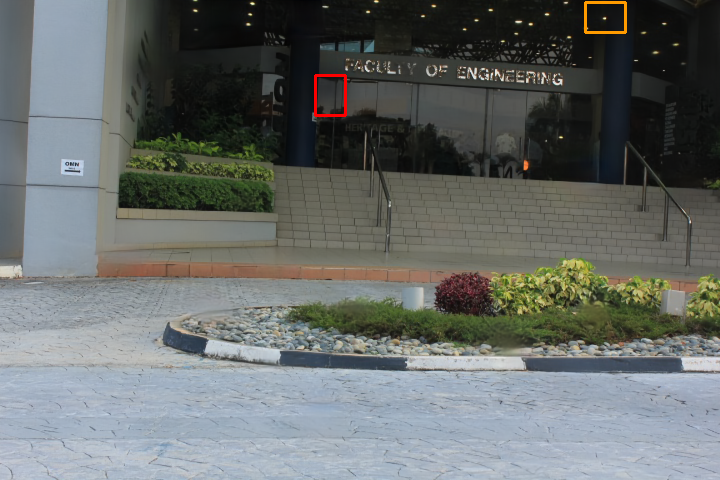}
    \end{subfigure}
    \hspace{-1mm}
     \begin{subfigure}{1\linewidth}
    \includegraphics[width=0.486\linewidth]{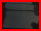}
    \hspace{-1mm}
     \includegraphics[width=0.486\linewidth]{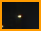}
    \end{subfigure}
    \subcaption[]{\cite{zhu2023learning_wgwsnet}}
    \label{fig:raindrop-2-d}
    \end{minipage}
    \hspace{-1.5mm}
  \begin{minipage}{0.1389\linewidth}
    \centering
    \begin{subfigure}{1\linewidth}
    \includegraphics[width=1\linewidth]{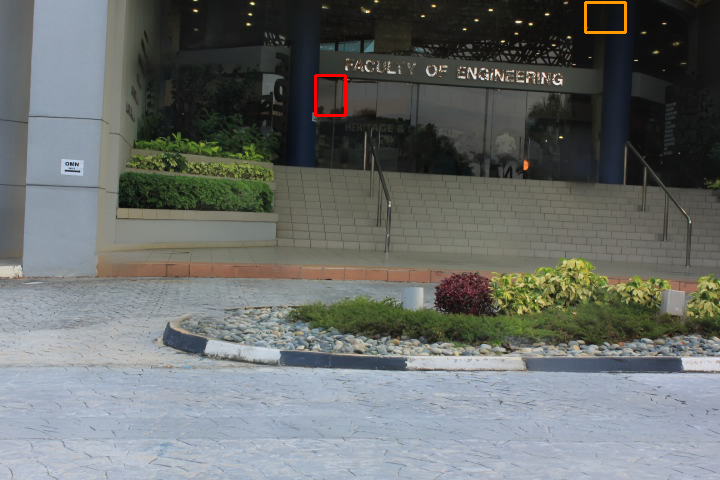}
    \end{subfigure}
    \hspace{-1mm}
     \begin{subfigure}{1\linewidth}
    \includegraphics[width=0.486\linewidth]{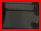}
    \hspace{-1mm}
     \includegraphics[width=0.486\linewidth]{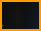}
    \end{subfigure}
    \subcaption[]{\cite{ozdenizci2023restoring}}
    \label{fig:raindrop-2-e}
    \end{minipage}
    \hspace{-1.5mm}
  \begin{minipage}{0.1389\linewidth}
    \centering
    \begin{subfigure}{1\linewidth}
    \includegraphics[width=1\linewidth]{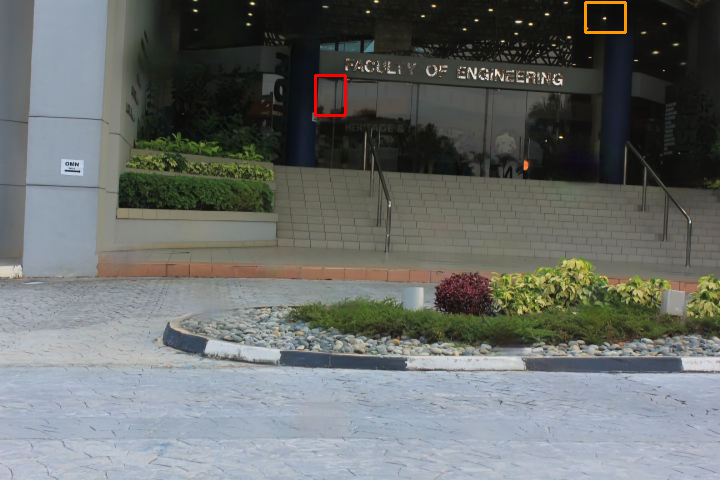}
    \end{subfigure}
    \hspace{-1mm}
     \begin{subfigure}{1\linewidth}
    \includegraphics[width=0.486\linewidth]{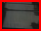}
    \hspace{-1mm}
     \includegraphics[width=0.486\linewidth]{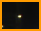}
    \end{subfigure}
    \subcaption[]{Ours}
    \label{fig:raindrop-2-f}
    \end{minipage}
    \hspace{-1.5mm}
  \begin{minipage}{0.1389\linewidth}
    \centering
    \begin{subfigure}{1\linewidth}
    \includegraphics[width=1\linewidth]{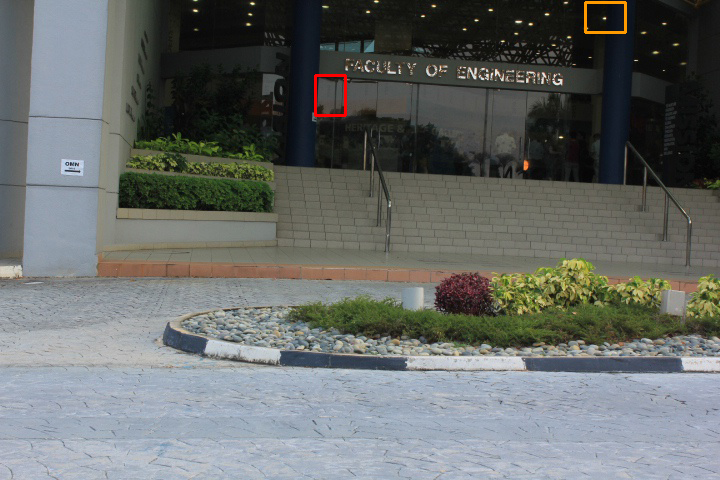}
    \end{subfigure}
    \hspace{-1mm}
     \begin{subfigure}{1\linewidth}
    \includegraphics[width=0.486\linewidth]{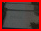}
    \hspace{-1mm}
     \includegraphics[width=0.486\linewidth]{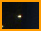}
    \end{subfigure}
    \subcaption[]{Clean}
    \label{fig:raindrop-2-g}
    \end{minipage}
  \caption{Visual comparison for raindrop removal on RainDrop~\cite{qian2018attentive}. The samples from (b) to (e) are Chen \textit{et al}.~\cite{Chen2022MultiWeatherRemoval}, TransWeather~\cite{valanarasu2022transweather}, WGWSNet~\cite{zhu2023learning_wgwsnet}, WeatherDiff~\cite{ozdenizci2023restoring}.}
  \label{fig:raindrop-2}
\end{figure*}
\begin{figure*}[t]
\setlength{\abovecaptionskip}{5pt}
\setlength{\belowcaptionskip}{0pt}
  \centering
  \begin{minipage}{0.1389\linewidth}
    \centering
    \begin{subfigure}{1\linewidth}
    \includegraphics[width=1\linewidth]{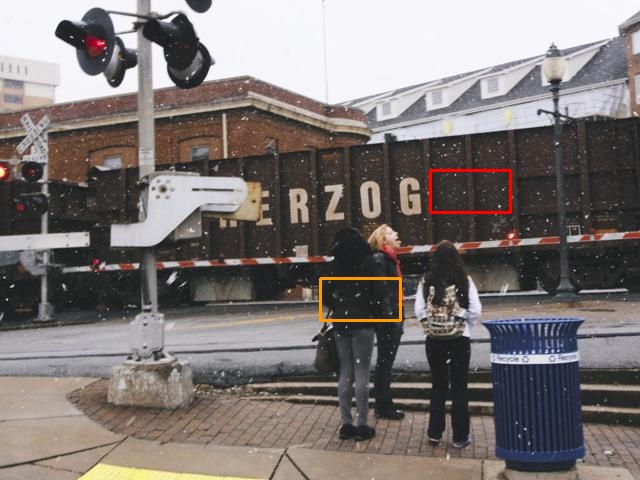}
    \end{subfigure}
    \hspace{-1mm}
     \begin{subfigure}{1\linewidth}
    \includegraphics[width=0.486\linewidth]{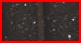}
    \hspace{-1mm}
    \includegraphics[width=0.486\linewidth]{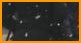}
    \end{subfigure}
    \subcaption[]{Input}
    \label{fig:realsnow-a}
    \end{minipage}
    \hspace{-1.5mm}
  \begin{minipage}{0.1389\linewidth}
    \centering
    \begin{subfigure}{1\linewidth}
    \includegraphics[width=1\linewidth]{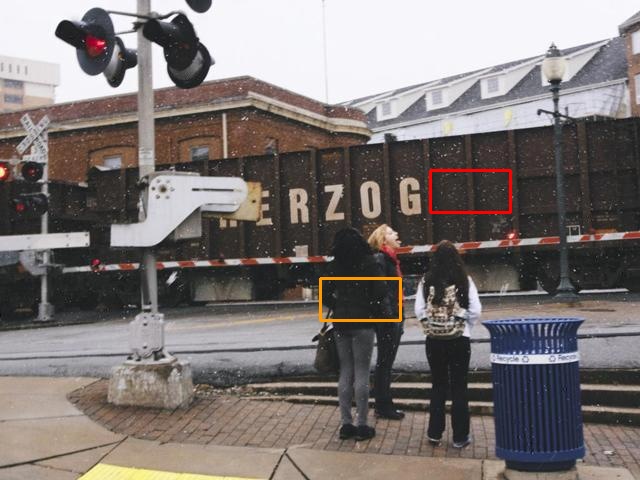}
    \end{subfigure}
    \hspace{-1mm}
     \begin{subfigure}{1\linewidth}
    \includegraphics[width=0.486\linewidth]{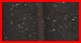}
    \hspace{-1mm}
    \includegraphics[width=0.486\linewidth]{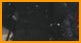}
    \end{subfigure}
    \subcaption[]{\cite{Chen2022MultiWeatherRemoval}}
    \label{fig:realsnow-b}
    \end{minipage}
    \hspace{-1.5mm}
  \begin{minipage}{0.1389\linewidth}
    \centering
    \begin{subfigure}{1\linewidth}
    \includegraphics[width=1\linewidth]{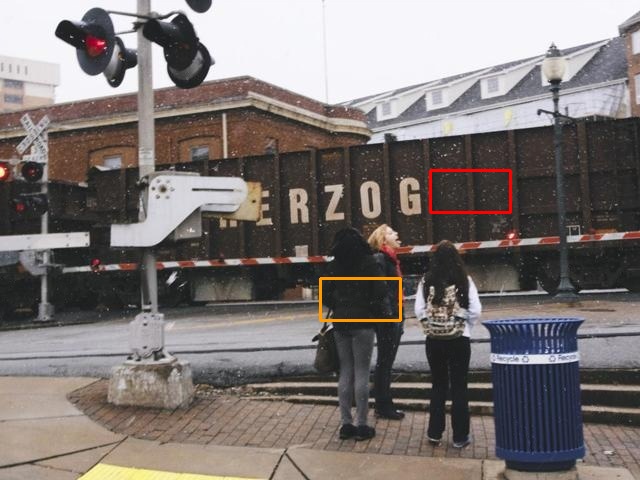}
    \end{subfigure}
    \hspace{-1mm}
     \begin{subfigure}{1\linewidth}
    \includegraphics[width=0.486\linewidth]{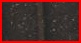}
    \hspace{-1mm}
    \includegraphics[width=0.486\linewidth]{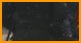}
    \end{subfigure}
    \subcaption[]{\cite{zamir2022restormer}}
    \label{fig:realsnow-c}
    \end{minipage}
    \hspace{-1.5mm}
  \begin{minipage}{0.1389\linewidth}
    \centering
    \begin{subfigure}{1\linewidth}
    \includegraphics[width=1\linewidth]{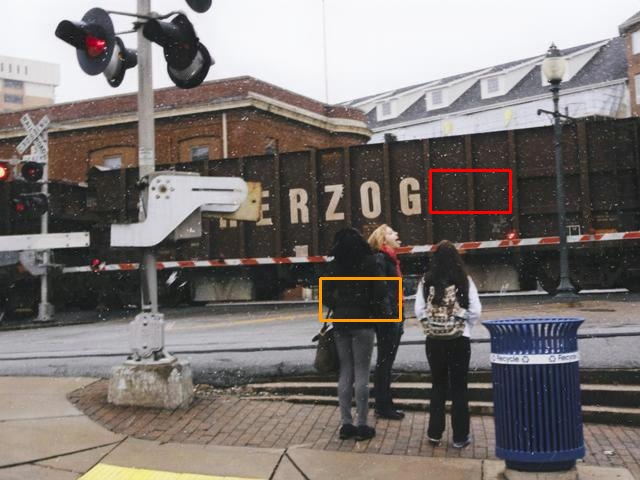}
    \end{subfigure}
    \hspace{-1mm}
     \begin{subfigure}{1\linewidth}
    \includegraphics[width=0.486\linewidth]{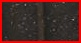}
    \hspace{-1mm}
    \includegraphics[width=0.486\linewidth]{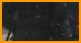}
    \end{subfigure}
    \subcaption[]{\cite{valanarasu2022transweather}}
    \label{fig:realsnow-d}
    \end{minipage}
    \hspace{-1.5mm}
  \begin{minipage}{0.1389\linewidth}
    \centering
    \begin{subfigure}{1\linewidth}
    \includegraphics[width=1\linewidth]{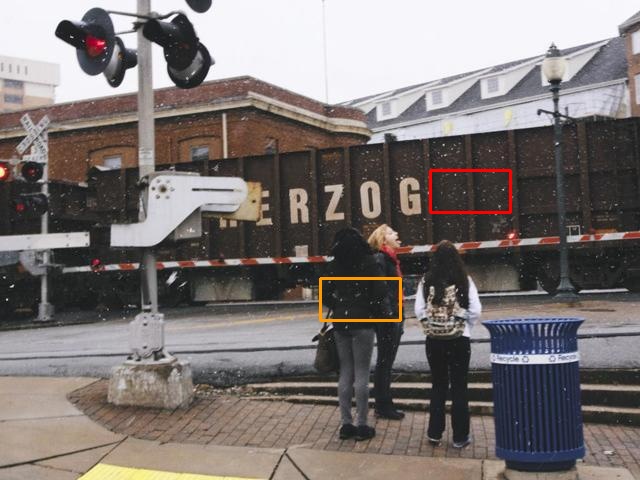}
    \end{subfigure}
    \hspace{-1mm}
     \begin{subfigure}{1\linewidth}
    \includegraphics[width=0.486\linewidth]{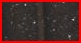}
    \hspace{-1mm}
    \includegraphics[width=0.486\linewidth]{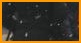}
    \end{subfigure}
    \subcaption[]{\cite{zhu2023learning_wgwsnet}}
    \label{fig:realsnow-e}
    \end{minipage}
    \hspace{-1.5mm}
  \begin{minipage}{0.1389\linewidth}
    \centering
    \begin{subfigure}{1\linewidth}
    \includegraphics[width=1\linewidth]{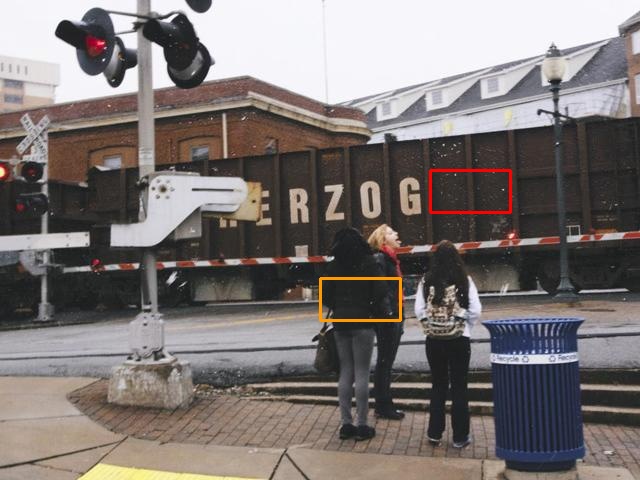}
    \end{subfigure}
    \hspace{-1mm}
     \begin{subfigure}{1\linewidth}
    \includegraphics[width=0.486\linewidth]{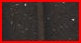}
    \hspace{-1mm}
    \includegraphics[width=0.486\linewidth]{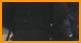}
    \end{subfigure}
    \subcaption[]{\cite{ozdenizci2023restoring}}
    \label{fig:realsnow-f}
    \end{minipage}
    \hspace{-1.5mm}
  \begin{minipage}{0.1389\linewidth}
    \centering
    \begin{subfigure}{1\linewidth}
    \includegraphics[width=1\linewidth]{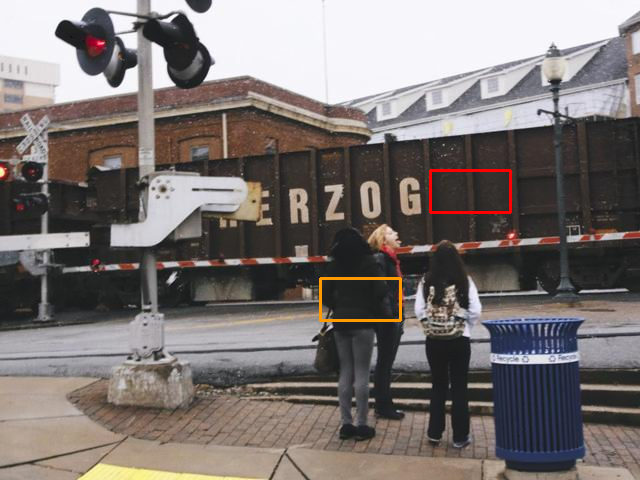}
    \end{subfigure}
    \hspace{-1mm}
     \begin{subfigure}{1\linewidth}
    \includegraphics[width=0.486\linewidth]{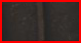}
    \hspace{-1mm}
    \includegraphics[width=0.486\linewidth]{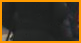}
    \end{subfigure}
    \subcaption[]{Ours}
    \label{fig:realsnow-g}
    \end{minipage}
  \caption{A qualitative comparison for real-world adverse weather removal on Snow100K~\cite{liu2018desnownet}. The samples from (b) to (e) are Chen \textit{et al}.~\cite{Chen2022MultiWeatherRemoval}, Restormer~\cite{zamir2022restormer}, TransWeather~\cite{valanarasu2022transweather}, WGWSNet~\cite{zhu2023learning_wgwsnet}, WeatherDiff~\cite{ozdenizci2023restoring}.}
  \label{fig:realsnow}
\end{figure*}

\section{Experiments}

\subsection{Experimental settings}

\paragraph{Datasets.}
We train our model on the same datasets as the previous works~\cite{li2020all,valanarasu2022transweather,ozdenizci2023restoring} to ensure a fair comparison.
The training set encompasses 9,000 images drawn from Snow100K~\cite{liu2018desnownet}, 1,069 images sourced from Raindrop~\cite{qian2018attentive}, and 9,000 images from Outdoor-Rain~\cite{li2019heavy}. 
Snow100K contains synthetic images deteriorated by snow, while Raindrop comprises real raindrop-affected images.
Outdoor-Rain features synthetic images afflicted by both fog and rain streaks. 
For evaluation, we employ the Test1 dataset~\cite{li2019heavy,li2020all}, the RainDrop test dataset~\cite{qian2018attentive}, and the Snow100K-L and -S test sets~\cite{liu2018desnownet}. 
Snow100K also provides a real-world test set containing 1,329 images affected by adverse weather.

\paragraph{Comparison Baselines.}
\looseness-1
We assess the performance of our method against state-of-the-art approaches designed specifically for distinct weather removal tasks: raindrop removal, snow removal, and rain\&fog removal. 
Specifically, for snow removal, we benchmark against SPANet~\cite{wang2019spatial}, JSTASR~\cite{chen2020jstasr}, RESCAN~\cite{li2018recurrent}, Desnow-Net~\cite{liu2018desnownet}, and DDMSNet~\cite{zhang2021deep}. 
In the case of rain\&fog removal, we compare with CycleGAN~\cite{zhu2017cyclegan}, pix2pix~\cite{isola2017pix2pix}, HRGAN~\cite{li2019heavy}, MPRNet~\cite{Zamir2021mprnet} and Restormer~\cite{zamir2022restormer}.
For raindrop removal, we evaluate against the methods such as pix2pix~\cite{isola2017pix2pix}, DuRN~\cite{liu2019durn}, RaindropAttn~\cite{quan2019deep}, AttentiveGAN~\cite{qian2018attentive}. 
Additionally, we include some recent transformer or multi-degradation restoration networks, IDT~\cite{xiao2022image}, NAFNet~\cite{chen2022simple}, MAXIM~\cite{tu2022maxim}, and Restormer~\cite{zamir2022restormer}, in our comparative analysis. 
It is worth noting that all these methods are single-task networks fine-tuned for specific datasets.

Furthermore, we conduct a performance comparison with the All-in-One network~\cite{li2020all}, Chen \textit{et al}.~\cite{Chen2022MultiWeatherRemoval}, TransWeather~\cite{valanarasu2022transweather}, WGWS-Net~\cite{zhu2023learning_wgwsnet}, WeatherDiff~\cite{ozdenizci2023restoring} and AWRCP~\cite{ye2023adverse}, which are trained to handle all the aforementioned tasks using a unified model. 
Note that our approach is also trained to tackle all these tasks using a single model.


\paragraph{Training details.}
Our implementation is realized by PyTorch~\cite{paszke2019pytorch} and on NVIDIA Tesla V100 GPU. 
The network is trained for a total of 300,000 iterations, with an initial batch size of 8 and an initial patch size of 128 akin to the progressive learning pipeline~\cite{zamir2022restormer}. 
We employ the AdamW optimizer~\cite{loshchilov2018adamw} with an initial learning rate of $3e^{-4}$ for the first $92,000$ iterations, which is gradually reduced to $1e^{-6}$ using cosine annealing schedule~\cite{loshchilov2016sgdr} during the remaining $208,000$ iterations. 
The number of blocks at each stage \(L_{i\in \{1,2,3,4\}}\) is set to \(\{4,4,6,8\}\) and the channel size \(C\) is 36. 
The channel expansion factor \(r\) in DGFF is set to $2.667$.
The numbers of heads in self-attention at different stages are set to $\{1,2,4,8\}$ respectively. 
We randomly apply horizontal and vertical flips as the technique of data augmentation. 


\subsection{Comparisons with the state-of-the-arts}
\label{sec:results}

\paragraph{Quantitative Evaluation.}

In our study, we provide a comprehensive comparative analysis of metrics applied to both synthetic and real datasets, as summarized in Table~\ref{tab:result}. 
For a fair and well-founded comparison, we utilize recent multiple degradation removal methods such as MPRNet~\cite{Zamir2021mprnet}, MAXIM~\cite{tu2022maxim}, and Restormer~\cite{zamir2022restormer}, treating them as weather-specific approaches for each benchmark.
Additionally, we retrain the all-in-one adverse weather removal methods including Chen \textit{et al}.~\cite{Chen2022MultiWeatherRemoval} and WGWS-Net~\cite{zhu2023learning_wgwsnet} using the all-weather training dataset~\cite{li2020all,valanarasu2022transweather,ozdenizci2023restoring}. 
This exhaustive comparison reveals that our proposed method exhibits a significant performance advantage over existing approaches across three different types of degradation.

\paragraph{Qualitative Evaluation.}
Furthermore, we conduct a visual comparison on three tasks, and the outcomes are showcased in Figure~\ref{fig:snow}, \ref{fig:rainfog} and \ref{fig:raindrop-2} respectively.
Figure~\ref{fig:realsnow} shows a case of real-world weather removal.
These results highlight that our method excels in comprehensively eliminating snow degradation, including fine and large snow spots. 
In contrast, the recent WeatherDiff~\cite{ozdenizci2023restoring} method still exhibits some residual snow degradation, and its capability to restore details is not optimal. 
When it comes to the restoration of challenging weather conditions, our method excels in removing complex haze and rain marks, yielding visually appealing results in comparison to prior approaches.

\subsection{Ablation studies}
\label{sec:ablation}

To substantiate the effectiveness of each component within Histoformer, we conduct a sequence of ablation studies on Outdoor-Rain~\cite{li2019heavy}. 
In particular, we examine the impact of 
the dynamic-range convolution, 
the DHSA module, 
the number of bins in DHSA, 
the DGFF module, 
and the correlation loss.

\paragraph{Dynamic-range Convolution.}
We experiment on two settings of dynamic-range convolution, namely, sorting horizontally first and then vertically before convolution, and the reverse order. 
Additionally, we compared them with vanilla convolution, and the results are displayed in Table~\ref{tab:abla_dynconv}. 
The operations of regular sorting led to a performance improvement of 0.14 dB, and the order of sorting operations does not significantly affect the outcome.

\paragraph{DHSA.}
To evaluate the effectiveness of the proposed DHSA module, we conduct a comparison with two baselines, i.e., a multi-Dconv head transposed attention (MDTA)~\cite{zamir2022restormer} and a top-k sparse attention (TKSA)~\cite{chen2023learning}. 
Additionally, we explore two additional settings of DHSA by excluding either the BHR branch or the FHR branch. 
The quantitative analysis results are presented in Table~\ref{tab:abla_sa}.

Both MDTA and TKSA integrate rich information across channels, which may result in a loss of the exploitation of long-range information across spatial dimensions. 
While our histogram self-attention can capture spatially long-range information, the use of either a single BHR or a single FHR branch neglects the inter-bin or inner-bin relationships, leading to inferior results.
By incorporating dynamic-range convolution and dual-branch histogram self-attention, capable of extracting long-range spatial features, our DHSA enhances performance, resulting in a PSNR improvement of 0.96 dB compared to TKSA.

\paragraph{Bins and Channels.}
To assess the influence of \(C\times B\), we conduct experiments with five different values on the first stage, i.e., $12$, $20$, $28$, $36$, and $44$. 
The results are presented in Table~\ref{tab:abla_bins}. 
It is observed that increasing the number of bins and channels consistently improves performance. 
However, when the number of \(C\times B\) exceeds $44$, it results in an out-of-memory error.

\begin{table}[tbp]
\setlength{\abovecaptionskip}{5pt}
\setlength{\belowcaptionskip}{0pt}
\tabcolsep=2mm
\scriptsize
  \centering
  \caption{Ablation studies on the dynamic-range convolution.}
    \begin{tabular}{cccc}
    \toprule
          & Vanilla Conv & Verti.+Horiz. & Horiz.+Verti. \\
    \midrule
    PSNR  & 31.94 & 32.03 & \textbf{32.08} \\
    SSIM  & 0.9377 & 0.9385 & \textbf{0.9389} \\
    \bottomrule
    \end{tabular}%
  \label{tab:abla_dynconv}%
\end{table}%

\begin{figure}[!t]
\setlength{\abovecaptionskip}{2pt}
\setlength{\belowcaptionskip}{0pt}
\begin{minipage}{\linewidth}
\scriptsize
\begin{minipage}[t]{0.487\linewidth}
\centering
\makeatletter\def\@captype{table}
\tabcolsep=2mm
    \caption{Ablation studies on the design of self-attention}\label{tab:abla_sa}%
    \begin{tabular}{lcc}
    \toprule
    SA type & PSNR  & SSIM \\
    \midrule
    MDTA~\cite{zamir2022restormer}  & 30.94 & 0.9278 \\
    TKSA~\cite{chen2023learning}  & 31.12 & 0.9295 \\
    w/o BHR & 31.05 & 0.9301 \\
    w/o FHR & 31.79 & 0.9364 \\
    DHSA  & \textbf{32.08} & \textbf{0.9389} \\
    \bottomrule
    \end{tabular}%
\end{minipage}
\hfill
\begin{minipage}[t]{0.487\linewidth}
\centering
\makeatletter\def\@captype{table}
\tabcolsep=2mm
    \caption{Ablation studies on the number of \(C\times B\)}\label{tab:abla_bins}%
    \begin{tabular}{ccc}
        \toprule
        \(C\times B\) & PSNR  & SSIM \\
        \midrule
        12    & 30.43 & 0.9283 \\
        20    & 31.56 & 0.9312 \\
        28    & 31.94 & 0.9379 \\
        36    & \textbf{32.08} & \textbf{0.9389} \\
        44    & \multicolumn{2}{c}{Out of memory} \\
        \bottomrule
    \end{tabular}%
\end{minipage}
\end{minipage}
\end{figure}

\begin{figure}[!t]
\setlength{\abovecaptionskip}{2pt}
\setlength{\belowcaptionskip}{0pt}
\begin{minipage}{\linewidth}
\scriptsize
\begin{minipage}[t]{0.487\linewidth}
\centering
\makeatletter\def\@captype{table}
\tabcolsep=2mm
    \caption{Ablation studies on the choice of feed-forward module}\label{tab:abla_ffn}%
    \begin{tabular}{lcc}
        \toprule
        Feed-Forward & PSNR  & SSIM \\
        \midrule
        FFN~\cite{liang2021swinir}    & 31.32 & 0.9313 \\
        GDFN~\cite{zamir2022restormer}   & 31.42 & 0.9347 \\
        DANB~\cite{zhao2023comprehensive}  & 31.56 & 0.9351 \\
        MSFN~\cite{chen2023learning}  & 31.78 & 0.9367 \\
        DGFF  & \textbf{32.08} & \textbf{0.9389} \\
        \bottomrule
    \end{tabular}%
\end{minipage}
\hfill
\begin{minipage}[t]{0.487\linewidth}
\centering
\makeatletter\def\@captype{table}
\renewcommand\arraystretch{0.89}
\tabcolsep=2mm
    \caption{Ablation studies on the setting of correlation loss}\label{tab:abla_corloss}%
    \begin{tabular}{llcc}
    \toprule
          & \(\alpha\) & PSNR  & SSIM \\
    \midrule
    w/o \(\mathcal{L}_{cor}\) & 0     & 31.77 & 0.9358 \\
    \midrule
    \multirow{4}[1]{*}{w/ \(\mathcal{L}_{cor}\)} & 0.1   & 32.01 & 0.9369 \\
          & 1     & \textbf{32.08} & \textbf{0.9389} \\
          & 5     & 32.03 & 0.9392 \\
          & 10    & 31.96 & 0.9375 \\
    \bottomrule
    \end{tabular}%
\end{minipage}
\end{minipage}
\end{figure}

\paragraph{DGFF.}
To assess the effectiveness of the proposed DGFF module, we conduct a comparison with four baselines: (i) the vanilla feed-forward network (FN)~\cite{liang2021swinir}, (ii) a gated-Dconv feed-forward network (GDFN)~\cite{zamir2022restormer}, (iii) a dual adaptive neural block (DANB)~\cite{zhao2023comprehensive}, and (iv) a mixed-scale
feed-forward network (MSFN)~\cite{chen2023learning}. 
The quantitative analysis results are presented in Table~\ref{tab:abla_ffn}. 
While MSFN integrates mixed-scale information, it may still miss out on the exploitation of multi-range spatial knowledge. 
Through the inclusion of pixel-shuffling and feature aggregation across different ranges, our DGFF further enhances performance, resulting in a PSNR gain of 0.3 dB over MSFN.

\paragraph{Correlation Loss.}
Table~\ref{tab:abla_corloss} shows the effectiveness of the correlation loss \(\mathcal{L}_{cor}\) and the influence of its weight. 
It is evident that \(\mathcal{L}_{cor}\) consistently improves the performance, while the specific loss weight does not have a substantial impact on the final results. 
We therefore keep the loss weight as $1$ by default.

\begin{figure}[t]
\setlength{\abovecaptionskip}{4pt}
\setlength{\belowcaptionskip}{0pt}
  \centering
  \begin{subfigure}{0.495\linewidth}
    \includegraphics[trim=35mm 81mm 35mm 52mm,clip,width=1\linewidth]{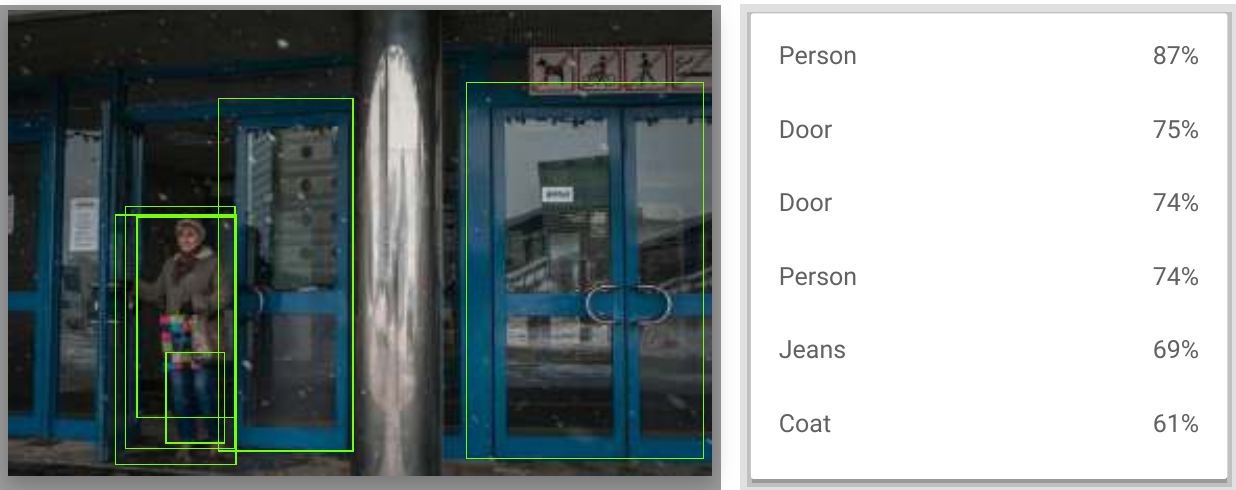}
  \end{subfigure}
  \hspace{-1mm}
  \begin{subfigure}{0.495\linewidth}
    \includegraphics[trim=35mm 81mm 35mm 52mm,clip,width=1\linewidth]{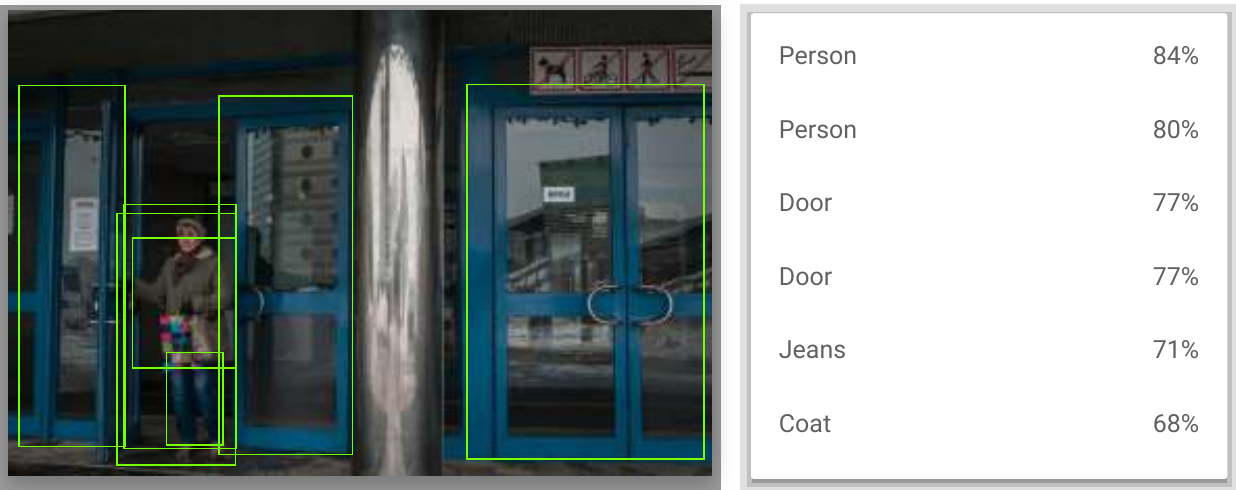}
  \end{subfigure}

    \centering
  \begin{subfigure}{0.495\linewidth}
    \includegraphics[trim=35mm 81mm 35mm 52mm,clip,width=1\linewidth]{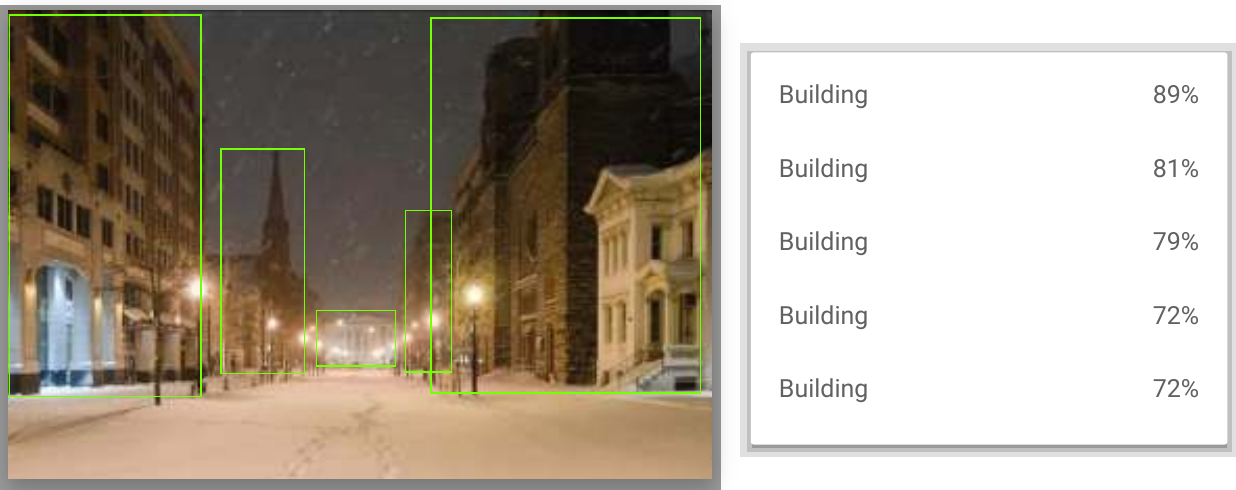}
    \caption{Input}
    \label{fig:real_detect-a}
  \end{subfigure}
  \hspace{-1mm}
  \begin{subfigure}{0.495\linewidth}
    \includegraphics[trim=35mm 81mm 35mm 52mm,clip,width=1\linewidth]{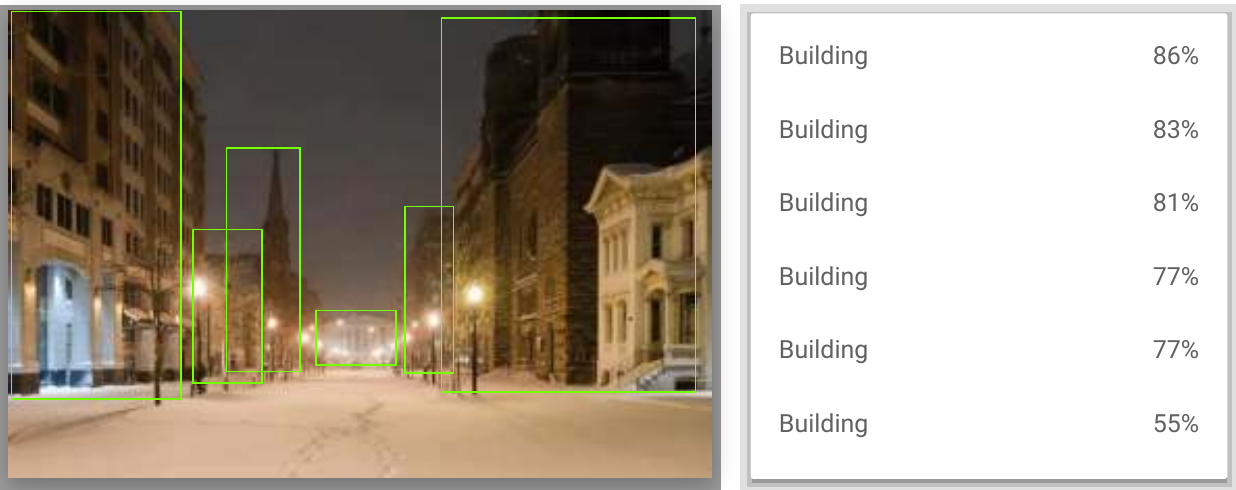}
    \caption{Deweathered by ours}
    \label{fig:real_detect-b}
  \end{subfigure}
  \caption{Real-world deweathering on two snowy images~\cite{liu2018desnownet} and their downstream detection results on \href{https://cloud.google.com/vision/docs/drag-and-drop}{Google API}.}
  \label{fig:real_detect}
\end{figure}
\subsection{Real-world Application}
To further demonstrate the practical applicability of our method for real-world adverse weather removal and its potential to improve downstream detection task, we provide two samples in Figure~\ref{fig:real_detect}. 
As depicted, our Histoformer effectively eliminates snowflakes from the scene and assists the detector in recognizing omitted door and building.


\section{Conclusion}
In this research, we introduce a novel mechanism called histogram self-attention and devise a new histogram transformer named Histoformer to tackle the challenge of all-in-one weather removal. 
Our histogram self-attention involves segmenting spatial features into multiple bins, and allocating varying attention along the bin or frequency dimension, allowing it to selectively focus on weather-related features with a dynamic range.
To facilitate learning both multi-range and multi-scale information, we present DGFF module and a correlation loss.
Through extensive experimentation, we demonstrate the effectiveness and superiority of our approach. 

\section*{Acknowledgement}
This work has been supported in part by National Natural Science Foundation of China (No. 62322216, 62172409, 62025604, 62306308, 62311530686), in part by Shenzhen Science and Technology Program (Grant No. JCYJ20220818102012025, KQTD20221101093559018), and in part by Guangdong Provincial Key Laboratory of Information Security Technology (No. 2023B1212060026).

{
    \bibliographystyle{splncs04}
    \bibliography{main}
}

\end{document}